\documentclass{article}

\usepackage[main,final]{neurips_2026}
\setcitestyle{numbers,square,comma}

\usepackage[utf8]{inputenc} 
\usepackage[T1]{fontenc}    
\usepackage[hidelinks]{hyperref}       
\usepackage{url}            
\usepackage{booktabs}       
\usepackage{amsfonts}       
\usepackage{nicefrac}       
\usepackage{microtype}      
\usepackage{xcolor}         

\usepackage{booktabs}

\usepackage{graphicx}
\usepackage{caption}
\usepackage{multirow}
\usepackage{booktabs}
\usepackage{wrapfig}
\usepackage{placeins}

\title{MIRA: Mutual Information guided calibration for Reliable Test-Time Adaptation}

\author{%
  \textbf{Hyeongyu Kim}$^{1,4}$%
  \thanks{These authors contributed equally to this work.}\hspace{0.8em}%
  \textbf{Mingyeong Jang}$^{1}$\footnotemark[1]\hspace{0.8em}%
  \textbf{Youngjun Song}$^{1}$\enspace
  \textbf{GeonHui Han}$^{1}$\enspace
  \textbf{Dosik Hwang}$^{1,2,3}$%
  \thanks{Corresponding author.}\\
  $^{1}$School of Electrical and Electronic Engineering,
  Yonsei University\\
  $^{2}$Department of Radiology and Center for Clinical
  Imaging Data Science,\\
  College of Medicine, Yonsei University\\
  $^{3}$Artificial Intelligence and Robotics Institute,
  Korea Institute of Science and Technology\\
  $^{4}$Hyundai Motor Company\\
  \texttt{\{lion4309, mini10211, dosik.hwang\}@yonsei.ac.kr}\\
  \textcolor{blue}{\url{https://hyeongyu-kim.github.io/mira/}}
}

\usepackage[table]{xcolor}
\usepackage{colortbl}
\usepackage{float}
\definecolor{lampblue}{HTML}{EAF3FF}
\definecolor{lampgreen}{HTML}{D8EFD3}
\usepackage{amsmath,amssymb}
\usepackage{needspace}

\usepackage{type1cm}
\usepackage{anyfontsize}
\DeclareFontShape{OT1}{cmr}{m}{n}{%
  <-5.5>cmr5 <5.5-7>cmr6 <7-8>cmr7 <8-9>cmr8
  <9-10>cmr9 <10-12>cmr10 <12-17>cmr12 <17->cmr17}{}
\DeclareFontShape{OML}{cmm}{m}{it}{%
  <-5.5>cmmi5 <5.5-7>cmmi6 <7-8>cmmi7 <8-9>cmmi8
  <9-10>cmmi9 <10-12>cmmi10 <12->cmmi12}{}

\newcommand{\MIRAtablefont}{\footnotesize}
\newcommand{\MIRAmethod}[1]{MIRA\smash{\textsubscript{#1}}}
\newcommand{\MIRAcompacttable}{%
  \MIRAtablefont
  \renewcommand{\arraystretch}{1.0}%
  \setlength{\aboverulesep}{0.8pt}%
  \setlength{\belowrulesep}{0.8pt}%
}
\DeclareMathSizes{8}{8}{6}{5}
\DeclareMathSizes{8.2065}{8.2065}{5.74455}{4.10325}
\newcommand{\MIRAlossfont}{\fontsize{8.2065}{9.5}\selectfont}
\newcommand{\MIRAsd}[1]{\kern0.2pt\hbox{\fontsize{3.8753}{4.5}\selectfont$\pm#1$}}

\begin{document}

\maketitle

\begin{abstract}
Test-time adaptation (TTA) updates a deployed model using only unlabeled target data, but standard entropy-based methods often rely on deterministic confidence and can reinforce overconfident errors under distribution shift.
We propose MIRA, \textbf{M}utual \textbf{I}nformation guided calibration for \textbf{R}eliable Test-Time \textbf{A}daptation, a source-free online adaptation method that uses lightweight last-layer Gaussian perturbations to probe prediction reliability.
For each target batch, MIRA reuses a single feature extraction pass and samples perturbed classifiers around the source classifier, forming an efficient local predictive distribution.
Rather than fitting a Bayesian posterior, MIRA treats the perturbation scale as a local probing radius and selects a batch-specific scale using mutual information (MI) and prediction agreement.
The resulting perturbation-sensitivity signal identifies predictions that are confident and locally stable under classifier perturbations, while suppressing brittle predictions with high stochastic disagreement.
MIRA then performs adaptation with an MI-confidence weighted entropy loss, which down-weights perturbation-sensitive or low-confidence samples before they drive entropy minimization.
This weighted loss reduces the influence of brittle predictions that can otherwise cause harmful adaptation under distribution shift.
The method requires no source data, retraining, model ensembles, or offline posterior fitting, and can be applied as a plug-and-play modification to entropy-minimization TTA.
Experiments on distribution-shift benchmarks show improvements in accuracy, calibration, and negative log-likelihood, suggesting that mutual-information-guided perturbation probing and weighted adaptation yield more robust online TTA than deterministic confidence.
\end{abstract}

\section{Introduction}

Deep neural networks perform well under i.i.d. training and test sampling from \(p(x,y)\) \cite{deng2009imagenet, he2016deep, krizhevsky2009learning}. 
Deployment often violates this assumption \cite{hendrycks2019benchmarking}. 
Under covariate shift, \(p_{\mathrm{src}}(x)\neq p_{\mathrm{tgt}}(x)\) with unchanged label space and task semantics \cite{shimodaira2000improving}. 
Such shifts, including corruptions from weather, sensor noise, styles, or acquisition conditions, degrade both accuracy and calibration \cite{ovadia2019can}. 
Confident errors are especially problematic.
Training-time domain generalization also seeks robustness to unseen domains, including through content--style disentanglement and style mixing~\cite{kim2023dimix}.

Complementing these training-time strategies, test-time adaptation (TTA) adapts a pre-trained model on unlabeled target data during inference, without access to source data or target labels \cite{sun2020test,tent,wang2025search,liang2025comprehensive,kim2026buffer, kim2026actta, song2026cd}. 
Existing methods improve robustness through normalization updates, entropy minimization, sample filtering, or consistency regularization \cite{tent, eata, cotta, zhang2022memo}. 

Despite this progress, most entropy-based TTA methods remain deterministic. 
They adapt from a single prediction of a single maximum a posteriori (MAP) network and often treat confidence as reliability. 
Under distribution shift, however, high confidence may reflect overconfidence rather than correctness \cite{guo2017calibration,ovadia2019can}. 
Entropy minimization explicitly sharpens target predictions \cite{amin2024over}; while useful when predictions are correct, it can also reinforce erroneous predictions and worsen overconfidence under severe shift \cite{amin2024over,zhang2024come}. Thus, reliable TTA requires more than confidence maximization.

Recent uncertainty-aware TTA methods estimate target reliability, filter updates, or incorporate stochasticity into adaptation \cite{bacs,lee2024aetta,tan2025uncertainty,brahma2023probabilistic,zhou2025bayesian}. 
However, many obtain reliability through mechanisms that complicate online deployment, such as ensembles \cite{bacs}, repeated stochastic forward passes \cite{lee2024aetta}, source-dependent posterior fitting \cite{bacs,brahma2023probabilistic}, uncertainty-specific regularizers \cite{tan2025uncertainty}, or oracle feedback with reinforcement-learning-based adaptation \cite{bitta}. 
PETAL fits a SWAG-diagonal posterior during source training and uses its log density to regularize lifelong teacher--student adaptation without revisiting source data \cite{brahma2023probabilistic}.
These costs conflict with the practical requirements of TTA: source-freeness, low overhead, immediacy, and plug-and-play deployment.

A related line of work studies efficient uncertainty estimation for neural networks \cite{llla,abnn}. 
Last-layer Laplace approximations place a lightweight distribution around the final classifier of a deterministic MAP network \cite{llla,daxberger2021laplace}, avoiding full Bayesian neural networks, ensembles, and full-network curvature estimation.
Post-hoc Bayesianization methods such as ABNN instead attach Bayesian adaptation layers to normalization layers and fine-tune pretrained models \cite{abnn}.
Yet these methods are typically offline post-hoc uncertainty estimators that rely on source data, validation data, or posterior fitting, making them unsuitable for online source-free TTA. 
At the same time, recent results such as ICLA show that useful uncertainty behavior can arise from simple isotropic approximations when the perturbation scale is chosen appropriately \cite{icla}. 
This suggests a more pragmatic question: can a lightweight perturbation scale be selected directly from the current unlabeled target batch, and can the resulting stochastic predictions be used to weight adaptation?

\begin{figure}[!t]
    \centering
    \includegraphics[width=0.99\linewidth]{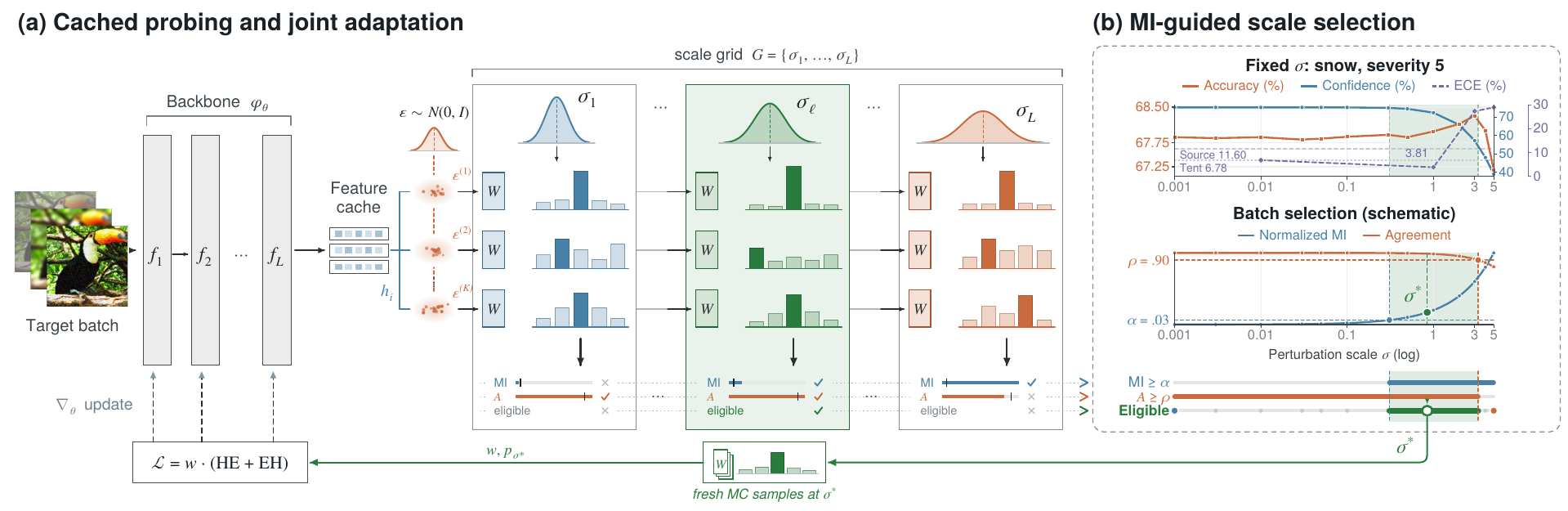}
    \caption{\textbf{MIRA overview.} (a) Cached features support Gaussian classifier probes with shared noise across candidate radii. At the selected $\sigma^*$, fresh Monte Carlo samples supply detached MI--confidence weights and a stochastic entropy update through one backward pass. (b) Fixed-scale annotations report measured results; MI/agreement curves and eligibility bars schematically illustrate radius selection. HE and EH denote entropy of the mean prediction and mean predictive entropy.}
\vspace{-5mm}
    \label{fig:main_results}
\end{figure}

We propose \textbf{MIRA}, a lightweight source-free online TTA framework that perturbs only the last-layer classifier of a deployed MAP network and uses stochastic predictions to identify reliable target samples. 
MIRA selects an informative perturbation scale per batch and applies an MI-confidence weighted entropy loss, promoting stable high-confidence predictions while down-weighting perturbation-sensitive samples. 
Figure~\ref{fig:main_results} summarizes the probe and reliability-weighted update. Because backbone features are computed once and reused across classifier samples, MIRA is efficient and plug-and-play. 
Across covariate-shift benchmarks, it improves accuracy, calibration, and NLL without source data, posterior fitting, ensembles, full-network stochastic inference, Hessian estimation, or retraining.

\noindent \textbf{Contributions.}\nopagebreak[4]

\noindent $\bullet$ We propose \textbf{MIRA}, a source-free online TTA framework that performs lightweight Gaussian perturbation probing around the last-layer classifier and selects a target-batch perturbation scale as an online reliability parameter, producing stochastic predictions useful for reliable adaptation and calibration without source data, validation data, or offline fitting.

\noindent $\bullet$ We present an MI-confidence weighted loss that treats MI as perturbation-induced disagreement and down-weights perturbation-sensitive or low-confidence samples before entropy minimization.

\noindent $\bullet$ We demonstrate consistent gains in accuracy and calibration across distribution-shift benchmarks, while keeping computation low through cached last-layer Monte Carlo sampling.

\section{Background}

\subsection{Test-Time Adaptation and Uncertainty}

Test-time adaptation (TTA) updates a source-trained model during inference using only unlabeled target samples~\cite{sun2020test,tent,liang2025comprehensive,wang2025search}. We consider the standard source-free covariate-shift setting, where \(p_{\mathrm{src}}(x) \neq p_{\mathrm{tgt}}(x)\) while the label space and task semantics remain unchanged. 

Given an incoming target batch \(\{x_i\}_{i=1}^{B}\), most TTA methods adapt the model by updating normalization statistics, minimizing prediction entropy, enforcing consistency, or selecting reliable samples for update~\cite{li2018adaptive,tent,eata,sar,zhang2022memo}. A representative objective is entropy minimization,
\begin{equation}
    \mathcal{L}_{\mathrm{ent}}
    =
    - \frac{1}{B}\sum_{i=1}^{B}\sum_{c=1}^{C}
    p_{\theta}(y=c \mid x_i)
    \log p_{\theta}(y=c \mid x_i),
\end{equation}
which encourages confident predictions on the target batch. 
While effective, this deterministic objective can treat confidence as reliability. 
Under severe shift, incorrect but low-entropy predictions may be reinforced, leading to overconfident or poorly calibrated adaptation.

Recent methods address this issue by incorporating uncertainty into TTA. 
EATA and SAR filter unreliable samples or unstable updates~\cite{eata,sar}; EATA-C models data and model uncertainty for calibration~\cite{tan2025uncertainty}; AETTA uses dropout-based disagreement to estimate adaptation quality without labels~\cite{lee2024aetta}; SSA stabilizes online adaptation through time-varying covariance~\cite{lee2025bayesian}; and BACS and PETAL introduce Bayesian or probabilistic formulations of adaptation~\cite{bacs,brahma2023probabilistic}. 
Uncertainty-constrained pseudo-label refinement also supports volumetric
self-training in cross-modality medical domain adaptation~\cite{shin2023sdcuda}.
Our goal is to obtain a lightweight predictive distribution directly from each
target batch and use it to guide adaptation, \textbf{without source data},
offline posterior fitting, or repeated full-network sampling.

\vspace{-1pt}
\subsection{Perturbation-Based Reliability Probing}

Predictive reliability can be probed through disagreement under stochastic predictors or controlled perturbations.
Monte Carlo dropout, deep ensembles, MEMO, and weight-space Bayesian methods capture uncertainty through stochastic subnetworks, independently trained models, input augmentations, or parameter distributions~\cite{gal2016dropout,lakshminarayanan2017simple,zhang2022memo,blundell2015weight,ovadia2019can}. These approaches may require repeated full-network inference, multiple models, source-side training, or additional fitting.
Self-supervised representation learning can structure features through disentanglement~\cite{lee2026fedi}, and pretrained feature spaces can support auxiliary self-supervision and consistency objectives~\cite{son2026hpgan}.
MIRA uses the deployed representation for reliability probing: it perturbs only the last-layer classifier and reuses cached backbone features within each batch, following cached-feature uncertainty estimators~\cite{daxberger2021laplace} and evidence that simplified perturbation scales can yield useful uncertainty behavior~\cite{icla}.
From the current unlabeled target batch, MIRA selects a Gaussian perturbation scale. MI measures perturbation-induced predictive disagreement, while prediction agreement checks class-identity stability. MIRA uses these perturbation responses to regulate sample-wise adaptation, directly linking predictive reliability to test-time optimization.

\vspace{-6pt}

\vspace{8pt}
\section{Motivation \& Observation}
\vspace{-8pt}

\begin{figure}[t]
    \centering
    \includegraphics[width=\linewidth]{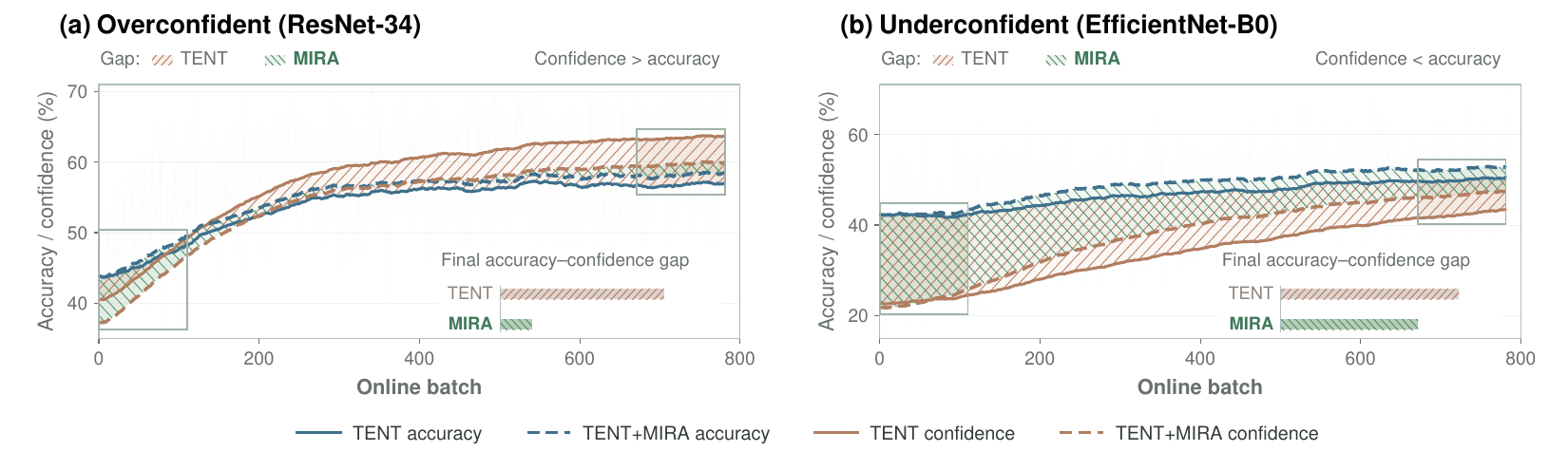}
    \caption{Online accuracy and confidence during TENT and TENT+MIRA on ImageNet-C fog (severity 5, EMA coefficient 0.01). MIRA narrows the gap in both (a) the overconfident ResNet-34 and (b) the underconfident EfficientNet-B0. Shading and bars show the absolute accuracy--confidence gap, rather than ECE.}
    \label{fig:online_calibration}
    \vspace{-8pt}
\end{figure}

Deterministic entropy-based TTA sharpens target predictions by minimizing the entropy of the model's own outputs.
This is effective when predictions are correct, but it treats low entropy as evidence of reliability without testing whether the resulting confidence is actually supported.
Consequently, confidence is not anchored to accuracy during adaptation.
Figure~\ref{fig:online_calibration} illustrates this on ImageNet-C fog: under TENT, confidence stays above accuracy for ResNet-34, whereas it stays below accuracy for EfficientNet-B0.
In both cases, the gap persists throughout the stream rather than closing as adaptation proceeds.

These observations motivate an additional reliability signal for adaptation.
Entropy minimization can reinforce the model's current prediction even when it is incorrect.
A single deterministic confidence score does not reveal whether that prediction is sensitive to small changes in the classifier.
We therefore ask whether prediction stability under controlled classifier perturbations can provide an additional signal for regulating adaptation.
\Needspace{5\baselineskip}
To examine this, we sweep fixed perturbation scales on CIFAR-100-C snow at severity 5 with ResNeXt (Figure~\ref{fig:main_results}(b)).
Starting from a deployed MAP classifier, we perturb only the last-layer weights,
\begin{equation}
    W_{\sigma}^{(k)}
    =
    W_{\mathrm{MAP}} + \sigma S\odot\epsilon_W^{(k)},
    \qquad
    \epsilon_W^{(k)}\sim\mathcal{N}(0,I).
\end{equation}
Here \(S\) has the same shape as \(W_{\mathrm{MAP}}\) and specifies the base perturbation scale for each weight; \(\odot\) denotes element-wise multiplication.
Section~4 specifies the corresponding weight and bias scale factors \(A_{\mathcal B}\) and \(a_{\mathcal B}\), with construction details in Appendix~\ref{app:scale}.
We sweep the perturbation scale \(\sigma\) on the incoming target stream.
Larger \(\sigma\) induces stronger classifier stochasticity.
This sweep is not intended to recover a Bayesian posterior; it probes how the target model responds as the classifier is gradually perturbed.

\noindent The sweep reveals two key observations.

\noindent \textbf{1. Perturbation scale can control confidence and calibration.}
When \(\sigma\) is very small, perturbed classifiers are nearly identical to the deterministic MAP model, so overconfident predictions remain unchanged.
As \(\sigma\) increases, perturbation averaging lowers confidence for predictions unstable under classifier perturbations.
In an intermediate regime, this reduces unsupported confidence while preserving most predictions, improving ECE and NLL.
If \(\sigma\) becomes too large, perturbed classifiers cross decision regions, causing confidence collapse and degraded probabilistic metrics.

\noindent \textbf{2. Perturbation search can also improve adaptation accuracy.}
The useful regime is not merely a confidence-smoothing region.
During TTA, the perturbation scale also changes the adaptation signal: brittle MAP predictions contribute less, while perturbation-stable predictions have greater influence.
This can suppress erroneous high-confidence updates and improve accuracy.
Thus, even without posterior fitting, searching a lightweight perturbation family can reveal scales that improve accuracy, calibration, or both.

These observations motivate MIRA. They suggest viewing \(\sigma\) not as a posterior variance but as a local probing radius around the deployed classifier.
A useful radius should be large enough to expose locally sensitive predictions, yet small enough to preserve class identity for stable samples; Appendix~\ref{app:app_A} formalizes this through local margin crossings.
Rather than fixing or pre-fitting the perturbation scale, MIRA searches each unlabeled target batch for this stable perturbation-response regime using MI and perturbation agreement, and uses the resulting stochastic predictions to compute MI--confidence weights for adaptation (Section~4).

\vspace{-8pt}

\Needspace{10\baselineskip}
\section{Method}
\vspace{-3pt}
MIRA performs source-free online test-time adaptation using lightweight
Gaussian perturbations of the final classifier.
Given a source-trained model, we decompose the prediction into a feature extractor
and a linear classifier,
\begin{equation}
    z(x) = W\phi_{\theta}(x) + b,
    \qquad
    p(y\mid x)=\mathrm{softmax}(z(x)),
\end{equation}
where \(\phi_{\theta}\) denotes the backbone representation and \(W,b\) denote
the last-layer classifier. MIRA updates only the parameters adapted by the
underlying TTA method, such as normalization affine parameters, while \(W,b\)
remain fixed. The perturbations serve two roles: they define a stochastic
adaptation objective and provide a reliability signal for weighting it.

\subsection{Last-Layer Gaussian Reliability Probe}

For an incoming target batch \(\mathcal{B}=\{x_i\}_{i=1}^{B}\), MIRA first
computes the backbone features once, \(h_i=\phi_{\theta}(x_i)\). It then samples
\(K\) perturbed last-layer classifiers for each perturbation scale \(\sigma\):
\begin{equation}
    W_{\sigma}^{(k)} = W + \sigma A_{\mathcal{B}}\odot \epsilon_W^{(k)},
    \qquad
    b_{\sigma}^{(k)} = b + \sigma a_{\mathcal{B}}\odot \epsilon_b^{(k)},
    \qquad
    \epsilon_W^{(k)},\epsilon_b^{(k)}\sim\mathcal{N}(0,I).
\end{equation}
Here \(A_{\mathcal{B}}\) and \(a_{\mathcal{B}}\) are fixed-rule scale factors
(see Appendix~\ref{app:scale}), so only the scalar radius \(\sigma\) is selected online.
Within each batch, the same noise \(\epsilon^{(k)}\) is shared across all scales, so candidate classifiers at different \(\sigma\) lie along the same perturbation directions and differ only in radius.
This coupling helps reduce Monte Carlo variability in comparisons across candidate scales.

For each sampled classifier, MIRA computes
\begin{equation}
    z_{i,\sigma}^{(k)} = W_{\sigma}^{(k)}h_i+b_{\sigma}^{(k)},\qquad
    p_{i,\sigma}^{(k)}=\mathrm{softmax}(z_{i,\sigma}^{(k)}),\qquad
    \bar p_{i,\sigma}=\frac{1}{K}\sum_{k=1}^{K}p_{i,\sigma}^{(k)}.
\end{equation}
From these stochastic predictions, MIRA computes predictive entropy, expected
entropy, and perturbation-induced mutual information:
\begin{equation}
    \mathrm{HE}_{i,\sigma}=H(\bar p_{i,\sigma}),\qquad
    \mathrm{EH}_{i,\sigma}=\frac{1}{K}\sum_{k=1}^{K}H(p_{i,\sigma}^{(k)}),\qquad
    \mathrm{MI}_{i,\sigma}=\mathrm{HE}_{i,\sigma}-\mathrm{EH}_{i,\sigma}.
\end{equation}
Here \(\mathrm{HE}\) measures uncertainty of the averaged stochastic prediction,
\(\mathrm{EH}\) measures the average uncertainty of individual sampled
classifiers, and \(\mathrm{MI}\) measures disagreement induced by perturbation.
MIRA also computes prediction agreement,
\begin{equation}
    a_{i,\sigma}
    =
    \frac{1}{K}\sum_{k=1}^{K}
    \mathbf{1}\!\left[
        \arg\max_y p_{i,\sigma}^{(k)}(y)
        =
        \arg\max_y \bar p_{i,\sigma}(y)
    \right],
\end{equation}
which checks whether sampled classifiers preserve the same class identity as the
Monte Carlo mean prediction.

\subsection{Mutual-Information-Guided Scale Selection}

The perturbation scale controls how strongly the local classifier neighborhood is
probed. As \(\sigma\) increases, sampled classifiers deviate more from the
deployed classifier and their predictions begin to disagree. A moderate rise in
mutual information can reveal local prediction sensitivity that deterministic
confidence misses. However, excessive disagreement is undesirable: when
\(\sigma\) is too large, sampled classifiers frequently cross decision regions,
and the resulting disagreement reflects predictive collapse rather than useful
sensitivity. MIRA therefore selects a batch-specific scale from
\(\mathcal{G}=\{\sigma_1,\ldots,\sigma_L\}\) by seeking disagreement that remains
class-consistent.

For each \(\sigma_\ell\in\mathcal{G}\), MIRA computes the batch-level mutual
information and prediction agreement,
\begin{equation}
    M(\sigma_\ell)=\frac{1}{B}\sum_{i=1}^{B}\mathrm{MI}_{i,\sigma_\ell},
    \qquad
    A(\sigma_\ell)=\frac{1}{B}\sum_{i=1}^{B}a_{i,\sigma_\ell}.
\end{equation}
Here \(M(\sigma_\ell)\) measures perturbation-induced disagreement and
\(A(\sigma_\ell)\) compares sampled top-1 classes with the Monte Carlo mean.
Let \(M_{\min}=\min_j M(\sigma_j)\) and
\(T_M=M_{\min}+\alpha[\max_j M(\sigma_j)-M_{\min}]\).
MIRA selects the earliest informative, high-agreement candidate:
\begin{equation}
    \sigma^{*}
    =\min_{\sigma_\ell\in\mathcal G}
    \left\{\sigma_\ell:M(\sigma_\ell)\ge T_M,\ A(\sigma_\ell)\ge\rho\right\}.
\end{equation}
The MI condition seeks an informative response relative to the observed MI range,
while the agreement condition checks whether the candidate remains class-consistent.
The complete selection procedure, including interpolation between grid points and
the case in which no candidate satisfies both conditions, is given in Appendix~D.6.
When both conditions hold, $1-A(\sigma^*)\le 1-\rho$ bounds the batch-average
empirical disagreement on the selection samples; it does not guarantee correctness,
preservation of the unperturbed MAP class, or agreement after fresh sampling.
The selected \(\sigma^*\) is a probing radius rather than a posterior variance.
Fresh Monte Carlo samples at this radius provide the adaptation loss below.

\subsection{Reliability-Gated Adaptation}

After selecting \(\sigma^*\), MIRA uses the stochastic predictions at this
scale to decide which target samples should drive adaptation. Standard entropy
minimization treats low entropy as reliable evidence and therefore reinforces the
model's current predictions. MIRA instead asks whether this confidence is stable
under controlled classifier perturbations. A sample with low entropy but high
perturbation-induced MI is locally brittle under the chosen perturbation
geometry: its confidence depends strongly on the particular classifier sample.
Such brittleness does not prove that the prediction is incorrect, but it
indicates that the sample should not dominate an entropy-minimization update.

Let
\begin{equation}
    \mathrm{HE}_i = H(\bar p_{i,\sigma^*}),\qquad
    \mathrm{EH}_i = \frac{1}{K}\sum_{k=1}^{K}H(p_{i,\sigma^*}^{(k)}),\qquad
    m_i=\mathrm{MI}_{i,\sigma^*}.
\end{equation}
MIRA first assigns each sample an MI-based weight that decreases with
perturbation-induced disagreement,
\begin{equation}
    w_i^{\mathrm{MI}}
    =
    \frac{\exp[-(m_i-\bar m)/\tau]}
    {B^{-1}\sum_{j=1}^{B}\exp[-(m_j-\bar m)/\tau]},
    \qquad
    \bar m = \frac{1}{B}\sum_{j=1}^{B}m_j .
\end{equation}
This normalization gives the weights unit batch mean. Thus, predictions that
remain stable across perturbed classifiers receive larger weights, while high-MI
predictions are suppressed. The centering by \(\bar m\) does not change the
relative weights, but stabilizes the batch-wise weighting scale. Unit-mean normalization redistributes average sample weight, but does not
by itself match gradient magnitudes; Appendix~\ref{app:matched} provides that control.

MIRA combines this MI gate with a confidence gate computed from the Monte Carlo
mean prediction,
\begin{equation}
    c_i=\max_y \bar p_{i,\sigma^*}(y),\qquad
    w_i^{\mathrm{conf}}
    =
    \mathrm{Norm}\!\left(c_i\right),
\end{equation}
where \(\mathrm{Norm}(\cdot)\) normalizes the batch mean weight to one. The final
sample weight is
\begin{equation}
    w_i
    =
    \mathrm{sg}\!\left[
    \mathrm{Norm}\!\left(
    w_i^{\mathrm{MI}} w_i^{\mathrm{conf}}
    \right)
    \right],
\end{equation}
where \(\mathrm{sg}[\cdot]\) stops gradients through the reliability estimate.
The MI term down-weights perturbation-sensitive samples, while the confidence
term down-weights low-evidence samples. Together, these factors favor samples
that are confident and low-disagreement under the selected probing radius.

The adaptation loss is
\begin{equation}
    \mathcal{L}_{\mathrm{MIRA}}
    =
    \frac{1}{B}\sum_{i=1}^{B}
    w_i\left(\mathrm{HE}_i+\gamma\mathrm{EH}_i\right).
\end{equation}
We use the common default \(\gamma=1\); coefficient sensitivity is reported
in Appendix~\ref{app:gamma}. The \(\mathrm{HE}\) term performs entropy minimization on the stochastic
predictive mean, and the \(\mathrm{EH}\) term encourages low entropy across the
sampled predictions. If \(w_i=1\), this reduces to the unweighted stochastic
objective \(\mathrm{HE}+\mathrm{EH}\), which provides the main adaptation signal;
the MI--confidence weight complements it by biasing the update away from
predictions that are sensitive to last-layer perturbations or have low confidence.
This does not guarantee pseudo-label correctness; it assumes that
perturbation-stable predictions are more likely to be correct, and reduces the
influence of brittle predictions that pose a higher risk of harmful
entropy-minimization updates under distribution shift.

\subsection{Efficient Cached Last-Layer Sampling}

MIRA is efficient because all Monte Carlo sampling is confined to the
last-layer reliability probe. Unlike dropout-based or full-network stochastic
methods, MIRA does not resample the backbone or perform repeated full forward
passes. For each target batch, it computes features once,
\(h_i=\phi_\theta(x_i)\), and reuses them for all classifier samples and all
candidate scales. The stochastic logits are computed only by final-layer
operations:
\begin{equation}
    z_{i,\sigma}^{(k)}
    =
    Wh_i+b
    +
    \sigma\big[
    (A_{\mathcal B}\odot\epsilon_W^{(k)})h_i
    +
    a_{\mathcal B}\odot\epsilon_b^{(k)}
    \big].
\end{equation}

MIRA selects a probing scale using the MI/agreement criteria (Appendix~D.6), then adapts with a reliability-weighted stochastic entropy objective using fresh Monte Carlo samples.
All sampling uses cached features and the final classifier, without source/validation data or posterior fitting.

\section{Experiments}

\vspace{-3pt}

We evaluate source-free online MIRA along three axes: accuracy and calibration,
MI-based reliability signals, and the efficiency of cached last-layer sampling.

\subsection{Experimental Setup}

\noindent{\textbf{Datasets and models.}}
We evaluate MIRA on CIFAR-100-C and ImageNet-C, each with 15 corruption types
and five severity levels. We use WideResNet-40-2 and ResNeXt on CIFAR-100-C,
and ResNet-34 and ResNet-50 on ImageNet-C. To assess generalization beyond
conventional CNN classifiers, we additionally evaluate ViT-B/16~\cite{dosovitskiy2021image} on ImageNet-C
and CLIP ViT-B/16~\cite{radford2021learning} on CIFAR-10-C and CIFAR-100-C. Unless otherwise specified,
adaptation is performed without access to source samples or target labels.

\noindent{\textbf{Baselines.}}
We compare against the source model without adaptation, TENT, EATA, SAR, and ROID~\cite{tent,eata,sar,roid}, and uncertainty-aware TTA methods in the extended comparisons. 
We also include perturbation- or uncertainty-estimation baselines when applicable to separate the effect of stochastic prediction from online adaptation.

\noindent{\textbf{Metrics.}}
We report classification accuracy for predictive performance and ECE and NLL for calibration and probabilistic prediction quality. 
We also report mean maximum confidence (MMC) to distinguish genuine calibration improvement from simple confidence sharpening. 
For reliability analysis, we compare batch accuracy at matched predictive entropy.

\noindent{\textbf{Implementation.}}
MIRA perturbs only the final classifier, using $K$ Monte Carlo samples per batch.
Cached backbone features are reused across samples and candidate scales, so
probing requires only last-layer operations. Unless otherwise specified, batch
size, optimizer, update parameters, and online protocol match each TTA baseline.
Appendix D provides baseline configurations, metric definitions, perturbation
construction, and hyperparameters.

\vspace{-3pt}

\subsection{Calibration with MI \& Agreement-Guided Scale Selection}
\vspace{-5pt}

\begin{figure}[htbp]
    \centering
    \includegraphics[width=1.00\linewidth]{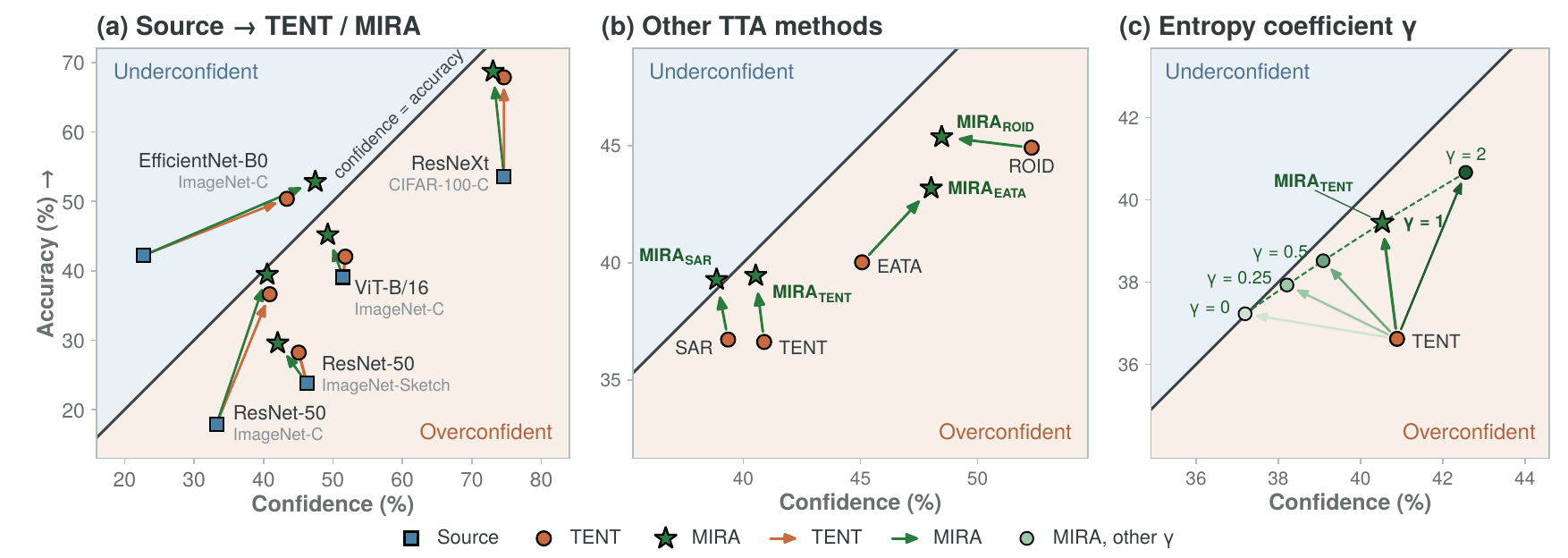}
    \caption{Confidence--accuracy plane; the diagonal marks confidence $=$ accuracy. (a) Architectures, (b) MIRA on base TTA methods, (c) $\gamma$ variants of MIRA$_{\text{TENT}}$ on ImageNet-C / ResNet-50.}
    
    \label{fig:fig3}
    \vspace{-4pt}
\end{figure}

Figure~\ref{fig:fig3} places adapted models in the confidence--accuracy plane. (a) Across architectures,
MIRA reaches higher accuracy than TENT while lying closer to the diagonal. (b) On ImageNet-C,
every base method is overconfident; MIRA raises accuracy and narrows the gap for all four.
(c) All $\gamma$ variants stay within 2 points of the diagonal and above TENT's accuracy.

We use MI and prediction agreement to select an informative probing scale and to check its class consistency. The earliest candidate satisfying both criteria is selected. This scale-selection role is separate from MI-based loss weighting, which we assess below. Additional reliability analyses are provided in Appendix~\ref{app:app_E}.

Joint $(\alpha,\rho)$, entropy-coefficient, and scale-grid sensitivity are reported
in Appendices~\ref{app:sensitivity}--\ref{app:grid}.
For CIFAR-100-C, accuracy spans only $0.16$ percentage points in the threshold
sweep, while ECE ranges from $0.0294$ to $0.0675$.
Coarsening the scale grid reduces ImageNet-C accuracy by $0.985$ points and
raises ECE by $0.0429$. Thus, the default is a common operating choice rather
than a universally optimal setting. Appendix~\ref{app:architecture_diagnostic} reports an $\mathrm{HE}$-only cross-architecture snow diagnostic.

\newsavebox{\losstablebox}
\sbox{\losstablebox}{\begingroup
\setlength{\tabcolsep}{0.7pt}
\renewcommand{\arraystretch}{1.0}
\newcommand{\lossentry}[2]{$#1\MIRAsd{#2}$}
\MIRAlossfont
\begin{tabular}{@{}lccccc@{}}
\toprule
\textbf{Method} & \textbf{Loss} & \textbf{Acc. $\uparrow$} & \textbf{MMC} & \textbf{ECE $\downarrow$} & \textbf{NLL $\downarrow$} \\
\midrule
Source & -- & \lossentry{0.1840}{0.0015} & \lossentry{0.3742}{0.0007} & \lossentry{0.1903}{0.0009} & \lossentry{5.4870}{0.0057} \\
TENT & \(H\) & \lossentry{0.3421}{0.0019} & \lossentry{0.4053}{0.0009} & \lossentry{0.0642}{0.0013} & \lossentry{3.4602}{0.0022} \\
TENT & \(2H\) & \lossentry{0.3660}{0.0021} & \lossentry{0.4388}{0.0012} & \lossentry{0.0734}{0.0012} & \lossentry{3.2892}{0.0026} \\
\midrule
\MIRAmethod{TENT} & \(\mathrm{HE}\) & \lossentry{0.3408}{0.0020} & \lossentry{0.3661}{0.0010} & \lossentry{0.0412}{0.0008} & \lossentry{3.4498}{0.0031} \\
\MIRAmethod{TENT} & \(\mathrm{EH}\) & \lossentry{0.3392}{0.0020} & \lossentry{0.3644}{0.0009} & \lossentry{0.0412}{0.0008} & \lossentry{3.4637}{0.0030} \\
\MIRAmethod{TENT} & \(\mathrm{MI}\) & \lossentry{0.2977}{0.0016} & \lossentry{0.3094}{0.0002} & \lossentry{\mathbf{0.0373}}{0.0003} & \lossentry{3.8173}{0.0075} \\
\MIRAmethod{TENT} & \(\mathrm{HE}+\mathrm{EH}\) & \lossentry{0.3637}{0.0024} & \lossentry{0.3961}{0.0013} & \lossentry{0.0442}{0.0011} & \lossentry{3.2806}{0.0055} \\
\midrule
\MIRAmethod{TENT} & \(w_{\mathrm{MI}}\mathrm{HE}\) & \lossentry{0.3425}{0.0021} & \lossentry{0.3675}{0.0010} & \lossentry{0.0415}{0.0007} & \lossentry{3.4387}{0.0021} \\
\MIRAmethod{TENT} & \(w_{\mathrm{MI}}(\mathrm{HE}+\mathrm{EH})\) & \lossentry{0.3658}{0.0023} & \lossentry{0.3971}{0.0014} & \lossentry{0.0441}{0.0010} & \lossentry{3.2681}{0.0007} \\
\rowcolor{lampgreen}
\MIRAmethod{TENT} & \(w_{\mathrm{MI}}w_{\mathrm{conf}}(\mathrm{HE}+\mathrm{EH})\) & \lossentry{\mathbf{0.3699}}{0.0024} & \lossentry{0.3992}{0.0011} & \lossentry{0.0435}{0.0020} & \lossentry{\mathbf{3.2421}}{0.0208} \\
\bottomrule
\end{tabular}
\endgroup

}
\vspace{-3pt}
\Needspace{\dimexpr\ht\losstablebox+\dp\losstablebox+80pt\relax}
\subsection{Loss Design: MI as a Reliability Gate}
\vspace{-3pt}
\begin{wraptable}{r}{\wd\losstablebox}
    \vspace{-8pt}
    \centering
    \caption{Loss ablation on ImageNet-C / ResNet-34: three-run mean $\pm$ sample SD (seeds 1, 2, and 3); accuracy is a fraction. Stochastic variants retain MI-based scale selection. Green: Full MIRA; bold: best means.}
    \label{tab:loss_ablation}
    \vspace{-6pt}
    \usebox{\losstablebox}
    \vspace{-6pt}
\end{wraptable}

MIRA provides the entropy of the Monte Carlo mean prediction $\mathrm{HE}$, the
expected entropy of sampled classifiers $\mathrm{EH}$, and their difference
$MI=\mathrm{HE}-\mathrm{EH}$. The entropy terms provide adaptation objectives: $\mathrm{HE}$ sharpens
the averaged stochastic prediction, while $\mathrm{EH}$ encourages individual perturbed
classifiers to remain confident. MI instead measures perturbation-induced
disagreement. Table~\ref{tab:loss_ablation} shows why it should not be the sole
optimization target: direct MI minimization lowers MMC and ECE but substantially
reduces accuracy and worsens NLL. A low ECE alone therefore does not imply a
better predictor. Neither $\mathrm{HE}$ nor $\mathrm{EH}$ alone improves accuracy over TENT,
whereas their combination increases accuracy to $36.37\%$
and reduces NLL to $3.2806$. This highlights the benefit
of jointly optimizing the mean prediction and individual
stochastic predictions.

The joint stochastic loss $\mathrm{HE}+\mathrm{EH}$ accounts for a substantial part of the
improvement over standard entropy minimization. MI weighting down-weights
perturbation-sensitive predictions, while the confidence gate suppresses
low-evidence samples. Applying $w_{\mathrm{MI}}$ to $\mathrm{HE}+\mathrm{EH}$ changes mean
accuracy from $36.37\%$ to $36.58\%$, with little change in ECE; Full MIRA
reaches $36.99\%$ and lower NLL. These modest differences should be interpreted
relative to seed variation and support gating as a complementary mechanism.
This ablation concerns loss weighting; MI-based scale selection
remains active in all stochastic variants. Gradient-matched controls are
reported in Appendix~\ref{app:matched}. Even under gradient-matched updates, Full MIRA improves accuracy
over unweighted $\mathrm{HE}+\mathrm{EH}$ by $0.28$ and $0.04$ percentage points
on CIFAR-10-C and CIFAR-100-C, respectively
(Appendix~\ref{app:matched}).

The deterministic $2H$ control shows why loss strength and calibration must
be assessed together. Doubling TENT's entropy objective raises accuracy to $36.60\%$,
but MMC increases to $0.4388$ and ECE to $0.0734$. Full MIRA instead reaches
$36.99\%$ with MMC $0.3992$ and ECE $0.0435$, so its accuracy gain accompanies
a smaller confidence--accuracy gap. This comparison supports evaluating
adaptation strength and calibration jointly. It does not isolate the gate by
itself, because stochastic prediction, scale selection, and weighting change
together; the gradient-matched comparisons in Appendix~\ref{app:matched}
address the separate question of sample weighting at a common update norm.
NLL complements this comparison by penalizing confident mistakes, whereas MMC
records the direction of confidence change and must be interpreted relative to accuracy.

\vspace{-3pt}

\newsavebox{\plugtablebox}
\sbox{\plugtablebox}{\begingroup
\setlength{\tabcolsep}{2.0pt}
\renewcommand{\arraystretch}{0.90}
\setlength{\aboverulesep}{0.8pt}
\setlength{\belowrulesep}{0.8pt}
\MIRAtablefont
\begin{tabular}{c|cccc|cccc}
\toprule
\multirow{2}{*}{\textbf{Method}} & \multicolumn{4}{c|}{\textbf{CIFAR-100-C / ResNeXt}} & \multicolumn{4}{c}{\textbf{ImageNet-C / ResNet-50}} \\
& Acc. (\%) $\uparrow$ & MMC & ECE $\downarrow$ & NLL $\downarrow$ & Acc. (\%) $\uparrow$ & MMC & ECE $\downarrow$ & NLL $\downarrow$ \\
\midrule
MAP & 53.55 & 0.7463 & 0.2109 & 2.4944 & 17.89 & 0.3322 & 0.1535 & 5.2161 \\
\midrule
TENT & $67.84\MIRAsd{0.05}$ & 0.7461 & 0.0693 & 1.2547 & $36.62\MIRAsd{0.12}$ & 0.4089 & 0.0478 & 3.2549 \\
\rowcolor{lampgreen}
\MIRAmethod{TENT} & $68.69\MIRAsd{0.15}$ & 0.7305 & 0.0452 & 1.1940 & $39.45\MIRAsd{0.14}$ & 0.4053 & 0.0361 & 3.0205 \\
\midrule
EATA & $65.37\MIRAsd{0.14}$ & 0.7039 & 0.0559 & 1.3515 & $40.02\MIRAsd{0.29}$ & 0.4507 & 0.0549 & 2.9796 \\
\rowcolor{lampgreen}
\MIRAmethod{EATA} & $68.47\MIRAsd{0.04}$ & 0.7217 & \textbf{0.0399} & 1.1921 & $43.18\MIRAsd{0.12}$ & 0.4802 & 0.0533 & 2.7547 \\
\midrule
SAR & $67.55\MIRAsd{0.11}$ & 0.7357 & 0.0630 & 1.2574 & $36.72\MIRAsd{0.15}$ & 0.3935 & 0.0355 & 3.2487 \\
\rowcolor{lampgreen}
\MIRAmethod{SAR} & $68.61\MIRAsd{0.10}$ & 0.7348 & 0.0500 & 1.2001 & $39.28\MIRAsd{0.56}$ & 0.3886 & \textbf{0.0290} & 3.0494 \\
\midrule
ROID & $69.16\MIRAsd{0.01}$ & 0.7644 & 0.0735 & 1.1931 & $44.91\MIRAsd{0.20}$ & 0.5231 & 0.0758 & 2.6526 \\
\rowcolor{lampgreen}
\MIRAmethod{ROID} & $\mathbf{69.21}\MIRAsd{0.12}$ & 0.7407 & 0.0497 & \textbf{1.1674} & $\mathbf{45.37}\MIRAsd{0.25}$ & 0.4847 & 0.0383 & \textbf{2.5916} \\
\specialrule{\heavyrulewidth}{0pt}{0pt}
\end{tabular}
\endgroup
}
\Needspace{\dimexpr\ht\plugtablebox+\dp\plugtablebox+80pt\relax}
\subsection{Calibration Plug-and-Play MI Gating}

\begin{wraptable}{r}{\wd\plugtablebox}
    \vspace{-8pt}
    \centering
    \caption{Plug-and-play evaluation: three-run means with accuracy $\pm$ SD. SDs for all metrics of \MIRAmethod{TENT} are reported in Table~\ref{tab:alpha_rho}. Bold: best accuracy, ECE, and NLL per dataset. Protocols: Appendix~\ref{app:setup}--\ref{app:metrics}.}
    \label{tab:plug_and_play_multi_model}
    \vspace{-6pt}
    \usebox{\plugtablebox}
    \vspace{-6pt}
\end{wraptable}

Beyond using MIRA as a standalone TTA method, we evaluate whether its mutual-information signal can be attached to existing adaptation rules. 
Many TTA methods differ in their sample-selection strategy or adaptation loss. 
MIRA provides an orthogonal reliability signal: low perturbation-induced MI indicates stable predictions, whereas high MI indicates brittle predictions that should be down-weighted. 
Thus, MIRA can be used as a lightweight plug-and-play gate without modifying the core update rule of the base method.
 
Table~\ref{tab:plug_and_play_multi_model} shows lower ECE and NLL with
method-dependent accuracy gains. Small accuracy differences, particularly ROID on
CIFAR-100-C, should be read relative to seed variation.
MIRA improves SAR's ImageNet-C accuracy from \(36.72\%\) to \(39.28\%\)
while reducing ECE and NLL.
 
The effect on confidence is not unidirectional, which is expected because the MI gate is not designed to monotonically sharpen or soften predictions.
MMC decreases for TENT, SAR, and ROID but increases for EATA on both benchmarks.
What is consistent is the absolute accuracy--confidence gap: every base method in Table~\ref{tab:plug_and_play_multi_model} is overconfident, and attaching MIRA narrows this gap in all eight settings.
For EATA, MMC rises, but accuracy rises even more, so the gap still shrinks.
This distinguishes MIRA from confidence-driven TTA, which typically pushes predictions to become sharper regardless of calibration.
By controlling which target samples influence adaptation, MIRA acts as a general reliability module for online TTA.

\vspace{-2pt}
\Needspace{185pt}
\subsection{Generalization and Comparison with Uncertainty Estimation Methods}

\newsavebox{\vittablebox}
\sbox{\vittablebox}{\begingroup
\setlength{\tabcolsep}{1.0pt}%
\renewcommand{\arraystretch}{1.0}%
\setlength{\aboverulesep}{0.8pt}%
\setlength{\belowrulesep}{0.8pt}%

\MIRAtablefont
\begin{tabular}{llrrr}
\toprule
\textbf{Dataset}
& \textbf{Method}
& \textbf{Acc. (\%) $\uparrow$}
& \textbf{ECE $\downarrow$}
& \textbf{NLL $\downarrow$} \\
\midrule

{ImageNet-C}
& TENT
& 42.03
& 0.1141
& 3.0312 \\

&
\cellcolor{lampgreen} \MIRAmethod{TENT}
& \cellcolor{lampgreen} \textbf{45.17}
& \cellcolor{lampgreen} \textbf{0.0529}
& \cellcolor{lampgreen} \textbf{2.8159} \\

\addlinespace[1pt]

{CIFAR-10-C}
& TENT
& 56.38
& 0.3629
& 3.5747 \\

&
\cellcolor{lampgreen} \MIRAmethod{TENT}
& \cellcolor{lampgreen} \textbf{58.65}
& \cellcolor{lampgreen} \textbf{0.3362}
& \cellcolor{lampgreen} \textbf{3.3509} \\

\addlinespace[1pt]

{CIFAR-100-C}
& TENT
& 31.28
& 0.4814
& 6.5500 \\

&
\cellcolor{lampgreen} \MIRAmethod{TENT}
& \cellcolor{lampgreen} \textbf{35.27}
& \cellcolor{lampgreen} \textbf{0.4066}
& \cellcolor{lampgreen} \textbf{5.5736} \\

\bottomrule
\end{tabular}\endgroup
}
\begin{wraptable}{r}{\wd\vittablebox}
    \vspace{-10pt}
    \centering
    \caption{ViT/CLIP generalization results.}
    \label{tab:vit_clip_generalization}
    \vspace{-6pt}
    \usebox{\vittablebox}
    \vspace{-8pt}
\end{wraptable}

\leavevmode\textbf{Generalization beyond conventional classifiers.}
Table~\ref{tab:vit_clip_generalization} extends MIRA to ViT-B/16 on ImageNet-C
and CLIP ViT-B/16 on CIFAR-10-C and CIFAR-100-C. Attaching MIRA to TENT improves
accuracy, ECE, and NLL in all three settings. For CLIP, class text embeddings
serve as the effective classifier and are perturbed only for reliability probing.
The released CLIPTTA’s TENT implementation ~\cite{lafon2025cliptta} uses ten updates per batch;
the modified one-step comparison and implementation details are given in
Appendices~\ref{app:clip_one_step} and~\ref{app:clip}.

\noindent{\textbf{Continual, dynamic, and computationally realistic adaptation.}}
MIRA also improves continual CoTTA on CIFAR-100-C from $64.16\%$ to $65.20\%$
accuracy, with ECE decreasing from $0.0529$ to $0.0342$.
Under a dynamic PTTA stream~\cite{yuan2023robust}, RoTTA improves from $74.89\%$ to $75.67\%$, with
ECE decreasing from $0.0842$ to $0.0674$.
Under the computationally realistic protocol, TENT accuracy improves from
$38.97\%$ to $39.92\%$ on ImageNet-C and from $40.74\%$ to $41.13\%$ on
ImageNet-R, with lower ECE and NLL on both.
These results and their full comparisons are given in
Appendix~\ref{app:app_C}, including the alternative PTTA protocol, where
accuracy and ECE improve but NLL worsens.

These extensions test accumulated adaptation, changing target distributions,
and time-limited update opportunities. Their accuracy--calibration trade-offs
complement the conventional corruption-stream comparison. The alternative
PTTA result illustrates why lower ECE alone does not establish better
probabilistic prediction: NLL worsens despite higher accuracy. We consequently
assess each protocol using accuracy, ECE, and NLL together.

\newsavebox{\bayestablebox}
\sbox{\bayestablebox}{\begingroup
\setlength{\tabcolsep}{3.0pt}
\renewcommand{\arraystretch}{0.95}
\setlength{\aboverulesep}{0.8pt}
\setlength{\belowrulesep}{0.8pt}

\MIRAtablefont
\begin{tabular}{@{}c|cccc|cccc}
\toprule

\multirow{3}{*}{\textbf{Method}}
& \multicolumn{4}{c|}{\textbf{CIFAR-100-C}}
& \multicolumn{4}{c}{\textbf{ImageNet-C}} \\
\cmidrule(lr){2-5} \cmidrule(lr){6-9}

& \multicolumn{4}{c|}{\textbf{WRN-40-2}}
& \multicolumn{4}{c}{\textbf{ResNet-50}} \\
\cmidrule(lr){2-5} \cmidrule(lr){6-9}

& \textbf{Acc $\uparrow$}
& \textbf{MMC}
& \textbf{ECE $\downarrow$}
& \textbf{NLL $\downarrow$}
& \textbf{Acc $\uparrow$}
& \textbf{MMC}
& \textbf{ECE $\downarrow$}
& \textbf{NLL $\downarrow$} \\

\midrule

MAP
& 0.5389 & 0.7041 & 0.1653 & 2.0589
& 0.2084 & 0.2246 & 0.0584 & 4.6490 \\

\midrule

Deep Ens.
& 0.5811 & 0.6269 & 0.0504 & 1.6998
& 0.2532 & 0.1874 & 0.0750 & 4.1725 \\

LLLA
& 0.5304 & 0.6282 & 0.0978 & 1.9300
& 0.2140 & 0.1948 & \textbf{0.0431} & 4.4985 \\

ICLA
& 0.5389 & 0.4449 & 0.0956 & 1.8801
& 0.2143 & 0.1879 & 0.0468 & 4.5034 \\

BBB
& 0.3445 & 0.5445 & 0.2001 & 3.7816
& 0.2057 & 0.3079 & 0.1035 & 4.8407 \\

ABNN
& 0.5179 & 0.6568 & 0.1389 & 2.2254
& 0.2326 & 0.2968 & 0.0634 & 4.2755 \\

\midrule

\rowcolor{lampgreen}
\MIRAmethod{TENT}
& \textbf{0.6391}
& 0.6701
& \textbf{0.0390}
& \textbf{1.3505}
& \textbf{0.3726}
& 0.3303
& 0.0512
& \textbf{3.2092} \\

\bottomrule
\end{tabular}
\endgroup}
\Needspace{\dimexpr\ht\bayestablebox+\dp\bayestablebox+65pt\relax}
\noindent\textbf{Comparison with uncertainty-estimation methods.}
We compare against deep ensembles, last-layer Laplace approximation (LLLA),
a simplified Laplace-style variant (ICLA), and Bayesian neural-network baselines
(BBB, ABNN) in Table~\ref{tab:table5_Bayes}.

\begin{wraptable}{r}{\wd\bayestablebox}
    \vspace{-8pt}
    \centering
    \caption{Uncertainty-estimation comparison. Accuracy is a fraction; values are retained single-run results.}
    \label{tab:table5_Bayes}
    \vspace{-6pt}
    \usebox{\bayestablebox}
    \vspace{-6pt}
\end{wraptable}

These baselines use uncertainty to form predictions, whereas MIRA uses local
perturbation responses to guide online adaptation. MIRA achieves the highest accuracy and lowest NLL on both benchmarks, and the lowest ECE on CIFAR-100-C. On ImageNet-C, Laplace-style methods attain slightly lower ECE,
but substantially lower accuracy and higher NLL.
Together with the loss ablation, these results support stochastic
adaptation complemented by reliability weighting while delimiting the gate's
incremental contribution.

\vspace{-2pt}
\Needspace{170pt}
\subsection{Efficiency of Last-Layer Probing}
\vspace{-2pt}

\newsavebox{\efficiencytablebox}
\sbox{\efficiencytablebox}{\setlength{\tabcolsep}{1.5pt}
\renewcommand{\arraystretch}{0.88}

\MIRAtablefont
\begin{tabular}{@{}ccccc@{}}
\toprule
\textbf{Method} &
\multicolumn{2}{c}{\shortstack{\textbf{CIFAR100}\\\textbf{WRN}}} &
\multicolumn{2}{c}{\shortstack{\textbf{ImageNet}\\\textbf{ResNet34}}} \\
\cmidrule(lr){2-3}
\cmidrule(lr){4-5}
&
\textbf{Runtime} &
\textbf{Memory} &
\textbf{Runtime} &
\textbf{Memory} \\
\midrule

Source
& \(1.00\times\) & \(1.00\times\)
& \(1.00\times\) & \(1.00\times\) \\

TENT
& \(1.48\times\) & \(1.68\times\)
& \(1.21\times\) & \(2.21\times\) \\

EATA
& \(1.67\times\) & \(1.68\times\)
& \(1.45\times\) & \(2.21\times\) \\

SAR
& \(1.90\times\) & \(1.45\times\)
& \(1.73\times\) & \(2.24\times\) \\

ROID
& \(2.81\times\) & \(2.06\times\)
& \(2.21\times\) & \(2.99\times\) \\

\rowcolor{lampgreen}
\textbf{\MIRAmethod{TENT}}
& \(2.05\times\) & \(2.05\times\)
& \(1.79\times\) & \(3.72\times\) \\

\bottomrule
\end{tabular}}
\begin{wraptable}{r}{\wd\efficiencytablebox}
    \vspace{-10pt}
    \centering
    \caption{Efficiency comparison. Runtime and memory are relative to the source model.}
    \label{tab:efficiency}
    \vspace{-6pt}
    \usebox{\efficiencytablebox}
    \vspace{-6pt}
\end{wraptable}

MIRA computes backbone features once per target batch and reuses them for all
Gaussian classifier samples and candidate scales. Thus, probing adds
classifier-level matrix multiplications rather than repeated backbone inference.
Table~\ref{tab:efficiency} reports moderate runtime overhead over TENT, with
runtime in the same range as SAR and ROID in the evaluated settings.
Memory can increase more: the ImageNet ratio is $3.72\times$ the source model,
compared with $2.21\times$ for TENT. Ablations on perturbation scope and the
Monte Carlo sampling budget are provided in Appendix B.

\vspace{-6pt}

\section{Conclusion}

We proposed MIRA, a source-free online TTA method that estimates reliability with last-layer Gaussian perturbations.
MIRA uses mutual information and prediction agreement, rather than a Bayesian posterior or deterministic confidence, to identify stable predictions.
By suppressing brittle high-confidence updates while preserving entropy-based adaptation, MIRA improves the accuracy--calibration trade-off; confining Monte Carlo sampling to the final classifier keeps it an efficient plug-and-play reliability module.

Across conventional classifiers, ViT, and CLIP, a local stochastic probe supported adaptation without repeated backbone evaluations.
Loss controls separate stochastic entropy from reliability weighting, and continual, dynamic, and time-constrained evaluations test the signal beyond conventional streams.
These findings suggest a role for MI alongside confidence: MI sets the probing scale, agreement checks class consistency, and sensitivity helps determine which predictions influence the update.
Future work should simplify scale selection and extend the probe to richer adaptation modules.

\noindent \textbf{Limitations.} MIRA adds a scale grid, thresholds \(\alpha,\rho\),
entropy coefficient \(\gamma\), and sampling budget \(K\).
Calibration remains sensitive to these choices and to grid density.
The agreement bound holds only when both selection conditions are met; under the default thresholds, the MI criterion alone often determines the scale. Stochastic entropy accounts for much of the accuracy gain,
and the gate's additional benefit is setting-dependent.
The probe is not calibrated Bayesian posterior uncertainty and cannot fully correct
severe feature-level misalignment. Adapting to unrepresentative unlabeled streams
may also reinforce existing bias, so confidence improvements do not establish
fairness or safe deployment across subpopulations.

\clearpage


\begin{ack}
This work was supported by the National Research Foundation of Korea (NRF) grant funded by the Korea government (MSIT) (No. RS-2025-16070382, RS-2025-02215070, RS-2025-02217919, RS-2024-00436680), Artificial Intelligence Graduate School Program at Yonsei University (RS 2020-II201361), the Korea Institute of Science and Technology (KIST) Institutional Program under Grant 26E0170.

\noindent\textbf{Competing Interests.} The authors declare no competing interests.
\end{ack}

{
\small

\bibliographystyle{plain} 
\bibliography{refs} 
}

\clearpage
\appendix
\raggedbottom
\setlength{\intextsep}{6pt}
\onecolumn
\section{Rationale for MIRA}
\label{app:app_A}

MIRA is based on the principle that predictions stable under controlled local
classifier perturbations provide safer signals for entropy-based adaptation than
locally brittle predictions. We do not interpret the resulting mutual information
as calibrated Bayesian posterior uncertainty; rather, the perturbation family is
used as an operational probe of local prediction stability.

To motivate the agreement criterion, consider the margin between the predicted
class $\hat y_i$ and a competing class $j$,
\begin{equation}
    m_{ij}=z_{i,\hat y_i}-z_{i,j}.
\end{equation}
Under Gaussian perturbation of the final classifier, the corresponding
pairwise boundary-crossing probability is
\begin{equation}
    \Pr[m_{ij}+\Delta m_{ij}<0]
    =
    \Phi\!\left(
        -\frac{m_{ij}}{\sigma s_{ij}}
    \right).
\end{equation}
Here $\hat y_i=\arg\max_c z_{i,c}$, $\Phi$ is the standard-normal CDF,
and $\Delta m_{ij}\sim\mathcal N(0,\sigma^2s_{ij}^2)$ conditional on the
cached features and perturbation factors. This motivates top-1 stability,
but MIRA measures empirical agreement with the Monte Carlo mean class;
it does not guarantee preservation of the unperturbed MAP class.

Mutual information provides complementary information. Agreement only changes
when the top-1 prediction crosses a decision boundary, whereas MI captures
changes in the full predictive distribution before such a crossing occurs.
Under a small-perturbation approximation,
\begin{equation}
    \mathrm{MI}_{i,\sigma}
    \approx
    \frac{1}{2}
    \operatorname{tr}\!\left(
        F(p_i)\Sigma_{z,i}
    \right).
\end{equation}
Here $p_i=\mathrm{softmax}(z_i)$,
$F(p_i)=\operatorname{diag}(p_i)-p_ip_i^\top$, and
$\Sigma_{z,i}=\operatorname{Cov}(\Delta z_i)$ includes the $\sigma^2$
scale factor. This gives perturbation-induced MI an interpretation as local
predictive sensitivity.

MIRA selects the earliest scale with non-trivial MI and high agreement,
balancing detectable predictive sensitivity with top-1 stability.
The entropy-matched analysis in Appendix~\ref{app:app_E} evaluates whether MI
provides reliability information beyond predictive entropy.

The exponential MI weight also follows entropy-regularized reliability
reweighting. For batch probabilities $q_i\ge0$ with $\sum_iq_i=1$,
\begin{equation}
 \min_q\sum_i q_i\mathrm{MI}_i+\tau\sum_iq_i\log q_i
 \quad\Longrightarrow\quad q_i\propto\exp(-\mathrm{MI}_i/\tau).
\end{equation}
The unit-mean weight is $w_i^{\mathrm{MI}}=Bq_i$; confidence weighting and
detachment then determine the final loss. Agreement provides its empirical
guard only when selection is feasible.

\section{Additional Ablation Studies}
\label{app:app_B}

New loss, gradient-control, construction, threshold, coefficient, and grid analyses
report mean $\pm$ sample SD across paired seeds 1, 2, and 3.
Accuracy SD is in percentage points; MMC, ECE, and NLL use their native units.
Each comparison retains its own backbone, optimizer, and stream protocol.
The existing scope and Monte Carlo-budget studies remain single-run.
Green highlights MIRA variants or the default configuration; bold marks the
best values in directional metrics within each comparison.

\subsection{Gradient-Matched Reliability Gating}
\label{app:matched}

Unit-mean sample weights do not ensure equal gradient norms.
To control uniform loss strengthening, we normalize the complete adaptation
gradient before the optimizer step.
We multiply each objective by a detached reciprocal of the complete adaptation
gradient's global L2 norm, targeting norm 1 before the optimizer step.
The original optimizer and learning rate are retained; the maximum recorded norm
error is below $2.49\times10^{-4}$.
CIFAR-10-C uses WRN AugMix, Adam at $10^{-3}$, $K=100$, and Gaussian noise, snow,
fog, and pixelation. CIFAR-100-C uses ResNeXt AugMix, SGD at $10^{-3}$,
$K=300$, and 15 corruptions. Both protocols reset between corruptions.

\begin{table}[H]
    \centering\small
    \caption{Gradient-matched loss and gating controls on CIFAR-10-C and CIFAR-100-C: global adaptation-gradient L2 norm 1, mean $\pm$ sample SD over three seeds. Panels use distinct protocols and are separate from ordinary adaptation. Bold marks best accuracy, ECE, and NLL within each panel.}
    \label{tab:gradient_matched}
    \begingroup
\setlength{\tabcolsep}{4pt}
\renewcommand{\arraystretch}{1.0}
\MIRAcompacttable
\begin{tabular}{@{}lrrrr@{}}
\toprule
Objective & Acc. (\%) $\uparrow$ & MMC & ECE $\downarrow$ & NLL $\downarrow$ \\
\midrule
\multicolumn{5}{l}{\textbf{CIFAR-10-C / WRN; Adam, four corruptions, $K=100$}} \\
\midrule
TENT & $87.79\MIRAsd{0.05}$ & $0.9412\MIRAsd{0.0005}$ & $0.0635\MIRAsd{0.0006}$ & $0.4462\MIRAsd{0.0025}$ \\
$\mathrm{HE}+\mathrm{EH}$ & $88.14\MIRAsd{0.05}$ & $0.9221\MIRAsd{0.0008}$ & $0.0409\MIRAsd{0.0011}$ & $0.3905\MIRAsd{0.0032}$ \\
+ MI gate & $88.34\MIRAsd{0.10}$ & $0.9236\MIRAsd{0.0007}$ & $0.0403\MIRAsd{0.0015}$ & $0.3838\MIRAsd{0.0027}$ \\
+ confidence gate & $88.29\MIRAsd{0.09}$ & $0.9229\MIRAsd{0.0008}$ & $0.0402\MIRAsd{0.0016}$ & $0.3848\MIRAsd{0.0025}$ \\
\rowcolor{lampgreen}
Full \MIRAmethod{TENT} & $\mathbf{88.42}\MIRAsd{0.07}$ & $0.9235\MIRAsd{0.0007}$ & $\mathbf{0.0396}\MIRAsd{0.0011}$ & $\mathbf{0.3808}\MIRAsd{0.0023}$ \\
\midrule
\multicolumn{5}{l}{\textbf{CIFAR-100-C / ResNeXt; SGD, $K=300$}} \\
\midrule
TENT & $65.97\MIRAsd{0.11}$ & $0.7150\MIRAsd{0.0004}$ & $0.0586\MIRAsd{0.0005}$ & $1.3250\MIRAsd{0.0029}$ \\
$\mathrm{HE}+\mathrm{EH}$ & $66.14\MIRAsd{0.10}$ & $0.6895\MIRAsd{0.0005}$ & $0.0333\MIRAsd{0.0005}$ & $1.2980\MIRAsd{0.0028}$ \\
+ MI gate & $\mathbf{66.27}\MIRAsd{0.12}$ & $0.6913\MIRAsd{0.0005}$ & $0.0333\MIRAsd{0.0005}$ & $\mathbf{1.2925}\MIRAsd{0.0033}$ \\
+ confidence gate & $66.11\MIRAsd{0.10}$ & $0.6872\MIRAsd{0.0005}$ & $0.0325\MIRAsd{0.0007}$ & $1.2979\MIRAsd{0.0023}$ \\
\rowcolor{lampgreen}
Full \MIRAmethod{TENT} & $66.18\MIRAsd{0.11}$ & $0.6880\MIRAsd{0.0005}$ & $\mathbf{0.0322}\MIRAsd{0.0009}$ & $1.2937\MIRAsd{0.0025}$ \\
\bottomrule
\end{tabular}
\endgroup

\end{table}

Relative to unweighted stochastic entropy, Full MIRA changes accuracy by
$+0.278$ and $+0.045$ percentage points on CIFAR-10-C and CIFAR-100-C,
respectively, with the same direction in all paired seeds.
These results support a setting-dependent contribution from reliability
weighting. Gradient normalization changes the optimizer step, so the absolute
accuracies in Table~\ref{tab:gradient_matched} are distinct from the main protocol.

\begin{table}[H]
    \centering\small
    \caption{Unnormalized CIFAR-10-C objective and gating controls, mean $\pm$ sample SD over three seeds. The Adam/four-corruption/$K=100$ protocol matches the corresponding normalized panel. Bold marks best accuracy, ECE, and NLL.}
    \label{tab:c10_vanilla}
    \begingroup
\setlength{\tabcolsep}{4pt}
\renewcommand{\arraystretch}{1.0}
\MIRAcompacttable
\begin{tabular}{@{}lrrrr@{}}
\toprule
Objective & Acc. (\%) $\uparrow$ & MMC & ECE $\downarrow$ & NLL $\downarrow$ \\
\midrule
TENT & $87.77\MIRAsd{0.06}$ & $0.9407\MIRAsd{0.0005}$ & $0.0634\MIRAsd{0.0008}$ & $0.4449\MIRAsd{0.0029}$ \\
$\mathrm{HE}+\mathrm{EH}$ & $88.15\MIRAsd{0.06}$ & $0.9218\MIRAsd{0.0007}$ & $0.0405\MIRAsd{0.0011}$ & $0.3902\MIRAsd{0.0038}$ \\
+ MI gate & $88.32\MIRAsd{0.04}$ & $0.9231\MIRAsd{0.0008}$ & $0.0401\MIRAsd{0.0011}$ & $0.3834\MIRAsd{0.0026}$ \\
+ confidence gate & $88.30\MIRAsd{0.10}$ & $0.9225\MIRAsd{0.0008}$ & $0.0398\MIRAsd{0.0016}$ & $0.3843\MIRAsd{0.0023}$ \\
\rowcolor{lampgreen}
Full \MIRAmethod{TENT} & $\mathbf{88.39}\MIRAsd{0.02}$ & $0.9232\MIRAsd{0.0007}$ & $\mathbf{0.0396}\MIRAsd{0.0004}$ & $\mathbf{0.3800}\MIRAsd{0.0029}$ \\
\bottomrule
\end{tabular}
\endgroup

\end{table}

\subsection{Perturbation Design Sensitivity}

We additionally examine whether MIRA is sensitive to the construction of the
last-layer perturbation scale. We separately vary the scale parameterization
and the base perturbation rule on CIFAR-100-C at severity level 5.
The WRN AugMix construction protocol uses all examples shuffled once, $K=100$,
SGD at $10^{-3}$, and reset between corruptions; it differs from the ResNeXt headline protocol.

\paragraph{Scale parameterization.}
We compare feature-wise, scalar, and class-wise parameterizations while keeping
the remaining MIRA configuration fixed.

\begin{table}[H]
    \centering
    \small
    \caption{Ablation on the perturbation-scale parameterization on CIFAR-100-C. Mean $\pm$ sample SD over seeds 1, 2, and 3; WRN AugMix, $K=100$, SGD $10^{-3}$. Bold marks best accuracy, ECE, and NLL.}
    \label{tab:supp_param}
    \begingroup
\setlength{\tabcolsep}{4pt}
\renewcommand{\arraystretch}{1.0}
\MIRAcompacttable
\begin{tabular}{@{}lrrrr@{}}
\toprule
Construction & Acc. (\%) $\uparrow$ & MMC & ECE $\downarrow$ & NLL $\downarrow$ \\
\midrule
TENT & $63.63\MIRAsd{0.08}$ & $0.7184\MIRAsd{0.0005}$ & $0.0824\MIRAsd{0.0008}$ & $1.4078\MIRAsd{0.0044}$ \\
\rowcolor{lampgreen}
Feature-wise & $\mathbf{64.33}\MIRAsd{0.10}$ & $0.6787\MIRAsd{0.0001}$ & $0.0387\MIRAsd{0.0009}$ & $1.3426\MIRAsd{0.0061}$ \\
Scalar & $64.29\MIRAsd{0.15}$ & $0.6589\MIRAsd{0.0006}$ & $\mathbf{0.0296}\MIRAsd{0.0007}$ & $1.3414\MIRAsd{0.0062}$ \\
Class-wise & $64.28\MIRAsd{0.15}$ & $0.6590\MIRAsd{0.0007}$ & $0.0297\MIRAsd{0.0008}$ & $\mathbf{1.3410}\MIRAsd{0.0062}$ \\
\bottomrule
\end{tabular}
\endgroup

\end{table}

The mean accuracy differences between feature-wise and scalar/class-wise
parameterizations are only $0.038/0.046$ percentage points; scalar and class-wise
variants have lower ECE. This indicates that MIRA is not highly sensitive to
the exact parameterization in accuracy, while calibration still depends on it.

\paragraph{Base perturbation scale.}
We next compare He-in, fan-in, weight-magnitude, and absolute base-scale rules.

\begin{table}[H]
    \centering
    \small
    \caption{Ablation on the base perturbation scale rule on CIFAR-100-C. Mean $\pm$ sample SD over seeds 1, 2, and 3; WRN AugMix, $K=100$, SGD $10^{-3}$. Bold marks best accuracy, ECE, and NLL.}
    \label{tab:supp_base}
    \begingroup
\setlength{\tabcolsep}{4pt}
\renewcommand{\arraystretch}{1.0}
\MIRAcompacttable
\begin{tabular}{@{}lrrrr@{}}
\toprule
Construction & Acc. (\%) $\uparrow$ & MMC & ECE $\downarrow$ & NLL $\downarrow$ \\
\midrule
TENT & $63.63\MIRAsd{0.08}$ & $0.7184\MIRAsd{0.0005}$ & $0.0824\MIRAsd{0.0008}$ & $1.4078\MIRAsd{0.0044}$ \\
\rowcolor{lampgreen}
He-in & $\mathbf{64.33}\MIRAsd{0.10}$ & $0.6787\MIRAsd{0.0001}$ & $0.0387\MIRAsd{0.0009}$ & $1.3426\MIRAsd{0.0061}$ \\
Fan-in & $64.27\MIRAsd{0.12}$ & $0.6599\MIRAsd{0.0001}$ & $0.0206\MIRAsd{0.0015}$ & $\mathbf{1.3326}\MIRAsd{0.0052}$ \\
Absolute & $64.19\MIRAsd{0.15}$ & $0.6440\MIRAsd{0.0001}$ & $\mathbf{0.0152}\MIRAsd{0.0006}$ & $1.3332\MIRAsd{0.0050}$ \\
Weight magnitude & $64.23\MIRAsd{0.13}$ & $0.6559\MIRAsd{0.0003}$ & $0.0246\MIRAsd{0.0009}$ & $1.3443\MIRAsd{0.0047}$ \\
\bottomrule
\end{tabular}
\endgroup

\end{table}

The different base-scale rules yield similar accuracy while all substantially
improve calibration over deterministic TENT. We therefore use the feature-wise
He-in construction as the fixed default rather than tuning the perturbation
rule for each dataset or model.

\subsection{Perturbation Scope}

We compare the default classifier-only perturbation with perturbing the last two
or three layers on ImageNet-C / ResNet-34 with batch size 64.

\begin{table}[H]
    \centering
    \small
    \caption{Effect of perturbation scope on ImageNet-C / ResNet-34 (batch size 64), retained single-run results. Bold marks the best values in directional columns.}
    \label{tab:perturbation_scope}
    \setlength{\tabcolsep}{5pt}
    \renewcommand{\arraystretch}{1.0}
    \MIRAcompacttable
\begin{tabular}{lccccccc}
        \toprule
        \textbf{Scope}
        & \textbf{Acc. (\%) $\uparrow$}
        & \textbf{MMC}
        & \textbf{ECE $\downarrow$}
        & \textbf{NLL $\downarrow$}
        & \textbf{Time (s) $\downarrow$}
        & \textbf{VRAM (MiB) $\downarrow$}
        & \textbf{Params} \\
        \midrule

        \rowcolor{lampgreen}
        Classifier only
        & 37.04 & 0.4013 & \textbf{0.0458} & 3.2690
        & \textbf{379.9} & \textbf{8180.4} & 513,000 \\

        Last 2 layers
        & 37.15 & 0.3576 & 0.0519 & 3.2372
        & 14,818.9 & 8212.3 & 2,872,296 \\

        Last 3 layers
        & \textbf{37.31} & 0.3592 & 0.0649 & \textbf{3.2332}
        & 21,572.5 & 8212.3 & 5,231,592 \\
        \bottomrule
    \end{tabular}
\end{table}

Deeper perturbations give only small accuracy/NLL gains while worsening ECE and
increasing runtime by roughly $39\times$--$57\times$. We therefore use the final
classifier as the best accuracy--calibration--efficiency trade-off.

\subsection{Monte Carlo Sampling Budget}

\begin{table}[H]
    \centering
    \small
    \caption{Effect of the Monte Carlo sampling budget $K$ on ImageNet-C / ResNet-50, retained single-run results. Bold marks the best values in directional columns.}
    \label{tab:k_budget}
    \setlength{\tabcolsep}{6pt}
    \renewcommand{\arraystretch}{1.0}
    \MIRAcompacttable
\begin{tabular}{rccccc}
        \toprule
        $K$
        & \textbf{Acc. (\%) $\uparrow$}
        & \textbf{ECE $\downarrow$}
        & \textbf{NLL $\downarrow$}
        & \textbf{Time (s) $\downarrow$}
        & \textbf{Peak VRAM (MiB) $\downarrow$} \\
        \midrule
        100  & 29.55 & 0.0410 & 3.6454 & \textbf{60} & \textbf{5,489} \\
        200  & 29.55 & 0.0408 & 3.6407 & 66 & 7,175 \\
        \rowcolor{lampgreen}
        300  & 29.39 & 0.0420 & \textbf{3.6374} & 74 & 8,671 \\
        500  & 29.29 & 0.0443 & 3.6511 & 88 & 11,939 \\
        1000 & \textbf{29.76} & \textbf{0.0381} & 3.6387 & 116 & 22,953 \\
        \bottomrule
    \end{tabular}
\end{table}

The scope and sampling-budget studies use different backbones, so their absolute
metrics should not be compared across tables. These retained single-run metrics
vary non-monotonically across $K$, while runtime
and memory increase with the sampling budget. We therefore use $K=300$ as a conservative
accuracy--efficiency trade-off.

\subsection{Joint MI and Agreement Threshold Sensitivity}
\label{app:sensitivity}

We evaluate $\alpha\in\{0.01,0.03,0.10\}$ and $\rho\in\{0.85,0.90,0.95\}$
jointly, retaining the Full objective, original 16-point grid, and $K=300$.
The panels use the headline TENT protocols: CIFAR-100-C / ResNeXt with SGD
$10^{-3}$ and ImageNet-C / ResNet-50 with SGD $2.5\times10^{-4}$.
Default controls shared with other sweeps are counted once.

\begin{table}[H]
    \centering\small
    \caption{Joint $\alpha$--$\rho$ sensitivity, mean $\pm$ sample SD over three seeds. The common default is $(0.03,0.90)$; test labels are used only for evaluation. Bold marks best accuracy, ECE, and NLL within each dataset; displayed ties are included.}
    \label{tab:alpha_rho}
    \begingroup
\setlength{\tabcolsep}{4pt}
\renewcommand{\arraystretch}{1.0}
\MIRAcompacttable
\begin{tabular}{@{}lrrrr@{}}
\toprule
Setting & Acc. (\%) $\uparrow$ & MMC & ECE $\downarrow$ & NLL $\downarrow$ \\
\midrule
\multicolumn{5}{l}{\textbf{CIFAR-100-C / ResNeXt}} \\
\midrule
$\alpha=0.01,\ \rho=0.85$ & $68.58\MIRAsd{0.11}$ & $0.7524\MIRAsd{0.0009}$ & $0.0675\MIRAsd{0.0010}$ & $1.2197\MIRAsd{0.0073}$ \\
$\alpha=0.01,\ \rho=0.90$ & $68.58\MIRAsd{0.12}$ & $0.7522\MIRAsd{0.0009}$ & $0.0675\MIRAsd{0.0012}$ & $1.2196\MIRAsd{0.0074}$ \\
$\alpha=0.01,\ \rho=0.95$ & $68.59\MIRAsd{0.13}$ & $0.7499\MIRAsd{0.0009}$ & $0.0651\MIRAsd{0.0010}$ & $1.2159\MIRAsd{0.0084}$ \\
$\alpha=0.03,\ \rho=0.85$ & $68.64\MIRAsd{0.13}$ & $0.7396\MIRAsd{0.0011}$ & $0.0542\MIRAsd{0.0009}$ & $1.2032\MIRAsd{0.0083}$ \\
\rowcolor{lampgreen}
$\alpha=0.03,\ \rho=0.90$ & $68.69\MIRAsd{0.15}$ & $0.7305\MIRAsd{0.0012}$ & $0.0452\MIRAsd{0.0007}$ & $1.1940\MIRAsd{0.0111}$ \\
$\alpha=0.03,\ \rho=0.95$ & $\mathbf{68.70}\MIRAsd{0.16}$ & $0.7279\MIRAsd{0.0011}$ & $0.0426\MIRAsd{0.0009}$ & $1.1906\MIRAsd{0.0113}$ \\
$\alpha=0.10,\ \rho=0.85$ & $68.55\MIRAsd{0.15}$ & $0.6576\MIRAsd{0.0011}$ & $\mathbf{0.0294}\MIRAsd{0.0003}$ & $\mathbf{1.1853}\MIRAsd{0.0094}$ \\
$\alpha=0.10,\ \rho=0.90$ & $68.54\MIRAsd{0.15}$ & $0.6570\MIRAsd{0.0011}$ & $0.0299\MIRAsd{0.0003}$ & $1.1859\MIRAsd{0.0093}$ \\
$\alpha=0.10,\ \rho=0.95$ & $68.54\MIRAsd{0.14}$ & $0.6570\MIRAsd{0.0011}$ & $0.0300\MIRAsd{0.0003}$ & $1.1859\MIRAsd{0.0093}$ \\
\midrule
\multicolumn{5}{l}{\textbf{ImageNet-C / ResNet-50}} \\
\midrule
$\alpha=0.01,\ \rho=0.85$ & $\mathbf{39.64}\MIRAsd{0.19}$ & $0.4435\MIRAsd{0.0018}$ & $0.0523\MIRAsd{0.0025}$ & $3.0261\MIRAsd{0.0118}$ \\
$\alpha=0.01,\ \rho=0.90$ & $\mathbf{39.64}\MIRAsd{0.19}$ & $0.4422\MIRAsd{0.0016}$ & $0.0514\MIRAsd{0.0028}$ & $3.0257\MIRAsd{0.0101}$ \\
$\alpha=0.01,\ \rho=0.95$ & $\mathbf{39.64}\MIRAsd{0.20}$ & $0.4419\MIRAsd{0.0016}$ & $0.0512\MIRAsd{0.0028}$ & $3.0253\MIRAsd{0.0111}$ \\
$\alpha=0.03,\ \rho=0.85$ & $39.44\MIRAsd{0.16}$ & $0.4053\MIRAsd{0.0014}$ & $0.0360\MIRAsd{0.0020}$ & $3.0216\MIRAsd{0.0022}$ \\
\rowcolor{lampgreen}
$\alpha=0.03,\ \rho=0.90$ & $39.45\MIRAsd{0.14}$ & $0.4053\MIRAsd{0.0014}$ & $0.0361\MIRAsd{0.0022}$ & $\mathbf{3.0205}\MIRAsd{0.0020}$ \\
$\alpha=0.03,\ \rho=0.95$ & $39.44\MIRAsd{0.16}$ & $0.4053\MIRAsd{0.0014}$ & $\mathbf{0.0359}\MIRAsd{0.0020}$ & $3.0206\MIRAsd{0.0022}$ \\
$\alpha=0.10,\ \rho=0.85$ & $38.47\MIRAsd{0.10}$ & $0.3058\MIRAsd{0.0008}$ & $0.0791\MIRAsd{0.0008}$ & $3.1142\MIRAsd{0.0029}$ \\
$\alpha=0.10,\ \rho=0.90$ & $38.43\MIRAsd{0.10}$ & $0.3058\MIRAsd{0.0008}$ & $0.0787\MIRAsd{0.0009}$ & $3.1150\MIRAsd{0.0036}$ \\
$\alpha=0.10,\ \rho=0.95$ & $38.44\MIRAsd{0.09}$ & $0.3058\MIRAsd{0.0007}$ & $0.0788\MIRAsd{0.0008}$ & $3.1147\MIRAsd{0.0023}$ \\
\bottomrule
\end{tabular}
\endgroup

\end{table}

At $\alpha=0.03$, ImageNet-C accuracy is nearly identical across $\rho$. In this
setting, the two conditions are jointly satisfied in at most 0.30\% of batches, so
the MI criterion determines $\sigma^*$ and $\rho$ has little effect; the flat
response reflects this rather than robustness to $\rho$. On CIFAR-100-C, where joint
satisfaction is more frequent, $\rho$ mainly affects calibration (Table~12).
Across both benchmarks, accuracy stays within 0.2 percentage points of the default
for $\alpha\in\{0.01, 0.03\}$ and all tested $\rho$, whereas calibration is more
sensitive to $\alpha$: $\alpha=0.10$ lowers CIFAR-100-C ECE but raises ImageNet-C ECE
to 0.079, and ImageNet-C ECE is lowest at $\alpha=0.03$. We therefore recommend
$\alpha\approx 0.03$ as the operating point and treat $\rho$ as a secondary setting.

\subsection{Entropy-Coefficient Sensitivity}
\label{app:gamma}

We vary only the $\mathrm{EH}$ coefficient in $w(\mathrm{HE}+\gamma \mathrm{EH})$ over
$\gamma\in\{0,0.25,0.5,1,2\}$, keeping the confidence exponent fixed at 1.
The selector, optimizer, learning rate, and ordinary gradient update remain
unchanged; $\gamma=1$ shares the Full baseline.

\begin{table}[H]
    \centering\small
    \caption{Entropy-coefficient sensitivity, mean $\pm$ sample SD over three seeds. The coefficient of EH changes independently of confidence weighting. Bold marks best accuracy, ECE, and NLL within each dataset.}
    \label{tab:gamma}
    \begingroup
\setlength{\tabcolsep}{4pt}
\renewcommand{\arraystretch}{1.0}
\MIRAcompacttable
\begin{tabular}{@{}lrrrr@{}}
\toprule
Setting & Acc. (\%) $\uparrow$ & MMC & ECE $\downarrow$ & NLL $\downarrow$ \\
\midrule
\multicolumn{5}{l}{\textbf{CIFAR-100-C / ResNeXt}} \\
\midrule
$\gamma=0$ & $67.91\MIRAsd{0.11}$ & $0.7144\MIRAsd{0.0006}$ & $\mathbf{0.0384}\MIRAsd{0.0003}$ & $1.2209\MIRAsd{0.0064}$ \\
$\gamma=0.25$ & $68.22\MIRAsd{0.11}$ & $0.7198\MIRAsd{0.0008}$ & $0.0402\MIRAsd{0.0007}$ & $1.2094\MIRAsd{0.0069}$ \\
$\gamma=0.5$ & $68.45\MIRAsd{0.12}$ & $0.7241\MIRAsd{0.0010}$ & $0.0420\MIRAsd{0.0006}$ & $1.2022\MIRAsd{0.0083}$ \\
\rowcolor{lampgreen}
$\gamma=1$ & $68.69\MIRAsd{0.15}$ & $0.7305\MIRAsd{0.0012}$ & $0.0452\MIRAsd{0.0007}$ & $\mathbf{1.1940}\MIRAsd{0.0111}$ \\
$\gamma=2$ & $\mathbf{68.87}\MIRAsd{0.11}$ & $0.7391\MIRAsd{0.0011}$ & $0.0517\MIRAsd{0.0009}$ & $1.1984\MIRAsd{0.0064}$ \\
\midrule
\multicolumn{5}{l}{\textbf{ImageNet-C / ResNet-50}} \\
\midrule
$\gamma=0$ & $37.23\MIRAsd{0.10}$ & $0.3719\MIRAsd{0.0012}$ & $\mathbf{0.0329}\MIRAsd{0.0013}$ & $3.1832\MIRAsd{0.0038}$ \\
$\gamma=0.25$ & $37.93\MIRAsd{0.14}$ & $0.3821\MIRAsd{0.0012}$ & $0.0331\MIRAsd{0.0015}$ & $3.1275\MIRAsd{0.0031}$ \\
$\gamma=0.5$ & $38.52\MIRAsd{0.17}$ & $0.3909\MIRAsd{0.0013}$ & $0.0343\MIRAsd{0.0017}$ & $3.0838\MIRAsd{0.0031}$ \\
\rowcolor{lampgreen}
$\gamma=1$ & $39.45\MIRAsd{0.14}$ & $0.4053\MIRAsd{0.0014}$ & $0.0361\MIRAsd{0.0022}$ & $3.0205\MIRAsd{0.0020}$ \\
$\gamma=2$ & $\mathbf{40.67}\MIRAsd{0.19}$ & $0.4256\MIRAsd{0.0016}$ & $0.0396\MIRAsd{0.0026}$ & $\mathbf{2.9442}\MIRAsd{0.0023}$ \\
\bottomrule
\end{tabular}
\endgroup

\end{table}

Increasing $\gamma$ can improve accuracy while worsening ECE. Equal
coefficients provide a common trade-off, rather than a unique metric-wise optimum.

\subsection{Scale-Grid Sensitivity}
\label{app:grid}

We keep $K=300$ fixed and compare the original grid with a coarser grid,
a denser grid, and an extended range (Appendix~\ref{app:selector}).
This tests scale coverage independently of the Monte Carlo-budget ablation.

\begin{table}[H]
    \centering\small
    \caption{Scale-grid sensitivity, mean $\pm$ sample SD over three seeds. Full MIRA and the headline dataset-specific TENT protocols are fixed. Bold marks best accuracy, ECE, and NLL within each dataset.}
    \label{tab:grid_metrics}
    \begingroup
\setlength{\tabcolsep}{4pt}
\renewcommand{\arraystretch}{1.0}
\MIRAcompacttable
\begin{tabular}{@{}lrrrr@{}}
\toprule
Grid & Acc. (\%) $\uparrow$ & MMC & ECE $\downarrow$ & NLL $\downarrow$ \\
\midrule
\multicolumn{5}{l}{\textbf{CIFAR-100-C / ResNeXt}} \\
\midrule
\rowcolor{lampgreen}
Default (16) & $\mathbf{68.69}\MIRAsd{0.15}$ & $0.7305\MIRAsd{0.0012}$ & $0.0452\MIRAsd{0.0007}$ & $1.1940\MIRAsd{0.0111}$ \\
Coarse (9) & $68.54\MIRAsd{0.15}$ & $0.6570\MIRAsd{0.0011}$ & $0.0299\MIRAsd{0.0004}$ & $1.1859\MIRAsd{0.0091}$ \\
Dense (31) & $68.68\MIRAsd{0.16}$ & $0.7317\MIRAsd{0.0011}$ & $0.0463\MIRAsd{0.0011}$ & $1.1938\MIRAsd{0.0115}$ \\
Extended (20) & $68.56\MIRAsd{0.17}$ & $0.6603\MIRAsd{0.0014}$ & $\mathbf{0.0273}\MIRAsd{0.0010}$ & $\mathbf{1.1838}\MIRAsd{0.0093}$ \\
\midrule
\multicolumn{5}{l}{\textbf{ImageNet-C / ResNet-50}} \\
\midrule
\rowcolor{lampgreen}
Default (16) & $39.45\MIRAsd{0.14}$ & $0.4053\MIRAsd{0.0014}$ & $\mathbf{0.0361}\MIRAsd{0.0022}$ & $3.0205\MIRAsd{0.0020}$ \\
Coarse (9) & $38.46\MIRAsd{0.10}$ & $0.3059\MIRAsd{0.0008}$ & $0.0789\MIRAsd{0.0010}$ & $3.1134\MIRAsd{0.0032}$ \\
Dense (31) & $\mathbf{39.59}\MIRAsd{0.15}$ & $0.4281\MIRAsd{0.0015}$ & $0.0429\MIRAsd{0.0030}$ & $3.0206\MIRAsd{0.0088}$ \\
Extended (20) & $39.46\MIRAsd{0.13}$ & $0.4045\MIRAsd{0.0014}$ & $0.0366\MIRAsd{0.0018}$ & $\mathbf{3.0196}\MIRAsd{0.0003}$ \\
\bottomrule
\end{tabular}
\endgroup

\end{table}

The default uses 16 candidate radii. Coarse9 reduces their density to 9;
Dense31 inserts geometric midpoints to obtain 31; Extended20 adds two smaller
and two larger radii (Appendix~\ref{app:selector}). These variants test scale
coverage and computational cost.
Coarse9 reduces ImageNet-C accuracy by $0.985$ percentage points and increases
ECE by $0.0429$; Dense31 increases accuracy by $0.142$ points, wall time from
$502.0$ to $583.7$ seconds, and steady allocated memory from $11.23$ to
$16.59$ GiB. Timing includes batch loading; steady memory excludes the first corruption.

No tested grid selects its boundary. Under the default thresholds, both conditions
are jointly satisfied in 22.63$\pm$0.95\% of CIFAR-100-C batches and in no
ImageNet-C batch; otherwise, $\sigma^*$ is the first candidate satisfying the MI
condition (Appendix~\ref{app:selector}). Because the MI threshold is defined
relative to the observed MI range, changing the grid also shifts the threshold.

\subsection{Cross-Architecture Numerical Values}
\label{app:architecture_diagnostic}

Table~\ref{tab:architecture_values} gives the values for the cross-architecture diagnostic:
ImageNet-C snow at severity 5, seed 1, 10,000 examples, $K=100$, and an
$\mathrm{HE}$ objective. The $(\alpha,\rho)$ settings are listed for each backbone.

\begin{table}[H]
    \centering\small
    \caption{Numerical values and selected settings for the cross-architecture diagnostic; single-run ImageNet-C snow diagnostics. Bold marks best accuracy, ECE, and NLL per backbone.}
    \label{tab:architecture_values}
    \begingroup
\setlength{\tabcolsep}{4pt}
\renewcommand{\arraystretch}{1.0}
\MIRAcompacttable
\begin{tabular}{@{}llrrrrrr@{}}
\toprule
Backbone & Method & $\alpha$ & $\rho$ & Acc. (\%) & MMC & ECE & NLL \\
\midrule
VGG-19-BN & Source & -- & -- & 13.24 & 0.3653 & 0.2329 & 5.5392 \\
VGG-19-BN & TENT & -- & -- & 34.60 & 0.4578 & 0.1118 & 3.4309 \\
\rowcolor{lampgreen}
VGG-19-BN & MIRA & 0.10 & 0.90 & \textbf{34.69} & 0.3359 & \textbf{0.0281} & \textbf{3.3540} \\
ResNet-50 & Source & -- & -- & 16.95 & 0.3945 & 0.2250 & 5.2586 \\
ResNet-50 & TENT & -- & -- & 39.55 & 0.4350 & 0.0395 & 2.9735 \\
\rowcolor{lampgreen}
ResNet-50 & MIRA & 0.01 & 0.90 & \textbf{39.67} & 0.4231 & \textbf{0.0266} & \textbf{2.9709} \\
ResNet-34 & Source & -- & -- & 15.47 & 0.4018 & 0.2471 & 5.5277 \\
ResNet-34 & TENT & -- & -- & 37.47 & 0.4299 & 0.0553 & 3.1236 \\
\rowcolor{lampgreen}
ResNet-34 & MIRA & 0.01 & 0.80 & \textbf{37.50} & 0.4171 & \textbf{0.0430} & \textbf{3.1211} \\
WRN-101-2 & Source & -- & -- & 22.24 & 0.4501 & 0.2277 & 4.7710 \\
WRN-101-2 & TENT & -- & -- & 44.78 & 0.4817 & 0.0423 & 2.7460 \\
\rowcolor{lampgreen}
WRN-101-2 & MIRA & 0.01 & 0.90 & \textbf{44.91} & 0.4709 & \textbf{0.0330} & \textbf{2.7411} \\
\bottomrule
\end{tabular}
\endgroup

\end{table}

\setlength{\intextsep}{6pt}      
\setlength{\textfloatsep}{6pt}   
\setlength{\floatsep}{6pt}       
\setlength{\abovecaptionskip}{3pt}
\setlength{\belowcaptionskip}{3pt}

\section{Additional Experiments}
\label{app:app_C}

We further evaluate MIRA under continual adaptation, alternative TTA protocols,
stronger distribution shifts, dynamic target streams, and computationally
realistic settings. Green marks MIRA variants; bold follows each table's
comparison scope.

\subsection{Continual and Dynamic Adaptation}

The continual comparison uses CIFAR-100-C / ResNeXt at severity 5.
The dynamic comparisons follow the PTTA protocol~\cite{yuan2023robust}, which
models changing distributions and correlated target samples.

\begin{table}[H]
\centering
\caption{Continual and dynamic adaptation. Bold compares each base/MIRA pair.}
\label{tab:continual_cifar100c}
\label{tab:dynamic_ptta}
\setlength{\tabcolsep}{4pt}
\MIRAcompacttable
\begin{tabular}{llrrrr}
\toprule
Protocol & Method & Acc. (\%) $\uparrow$ & MMC & ECE $\downarrow$ & NLL $\downarrow$ \\
\midrule
Continual & CoTTA & 64.16 & 0.6170 & 0.0529 & 1.4173 \\
\rowcolor{lampgreen}
& \MIRAmethod{CoTTA} & \textbf{65.20} & 0.6787 & \textbf{0.0342} & \textbf{1.3459} \\
\midrule
Dynamic & TENT & 14.75 & 0.9675 & 0.8201 & 17.6335 \\
\rowcolor{lampgreen}
& \MIRAmethod{TENT} & \textbf{15.16} & 0.9650 & \textbf{0.8134} & \textbf{17.0462} \\
\midrule
Dynamic & RoTTA & 74.89 & 0.8331 & 0.0842 & 0.8933 \\
\rowcolor{lampgreen}
& \MIRAmethod{RoTTA} & \textbf{75.67} & 0.8241 & \textbf{0.0674} & \textbf{0.8622} \\
\bottomrule
\end{tabular}
\end{table}

MIRA improves accuracy, ECE, and NLL for all three base-method pairs.
A separate reset-based ImageNet-C / ResNet-50 comparison under the PTTA protocol ( batch size 64, 10,000 examples) increases accuracy
from 37.50\% to 40.67\% and decreases ECE from 0.0284 to 0.0278, while NLL
increases from 2.7080 to 2.9286.

\subsection{Computationally Realistic Adaptation}

We additionally evaluate MIRA on ImageNet-C and ImageNet-R~\cite{hendrycks2021many} using ResNet-50
and batch size 64 under the time-constrained protocol of
Alfarra et al.~\cite{alfarra2024evaluation}. The stream arrival rate is
$\eta r$, where $r$ is the forward-pass rate of the non-adapted model.
The recorded conventional and realistic settings use $\eta=0$ and $\eta=1$,
respectively. In the conventional setting, every batch can be used for
adaptation regardless of processing time. In the realistic setting, batches
arrive at a constant rate, so a slower method receives fewer adaptation opportunities.

\begin{table}[H]
    \centering
    \small
    \caption{
    Evaluation under conventional ($\eta=0$) and computationally realistic
    ($\eta=1$) TTA protocols.
    Bold compares each dataset and realistic regime.
    }
    \label{tab:realistic_protocol}
    \setlength{\tabcolsep}{5pt}
    \renewcommand{\arraystretch}{1.0}
    \MIRAcompacttable
    \begin{tabular}{llccccc}
        \toprule
        \textbf{Dataset}
        & \textbf{Method}
        & \textbf{Realistic}
        & \textbf{Acc. (\%) $\uparrow$}
        & \textbf{MMC}
        & \textbf{ECE $\downarrow$}
        & \textbf{NLL $\downarrow$} \\
        \midrule

        \multirow{4}{*}{ImageNet-C}
        & TENT
        & No
        & 42.71 & 0.5203 & 0.0935 & 2.9071 \\

        & \cellcolor{lampgreen}\MIRAmethod{TENT}
        & \cellcolor{lampgreen}No
        & \cellcolor{lampgreen}\textbf{45.02}
        & \cellcolor{lampgreen}0.5007
        & \cellcolor{lampgreen}\textbf{0.0573}
        & \cellcolor{lampgreen}\textbf{2.7948} \\

        & TENT
        & Yes
        & 38.97 & 0.4444 & 0.0572 & 3.0884 \\

        & \cellcolor{lampgreen}\MIRAmethod{TENT}
        & \cellcolor{lampgreen}Yes
        & \cellcolor{lampgreen}\textbf{39.92}
        & \cellcolor{lampgreen}0.4412
        & \cellcolor{lampgreen}\textbf{0.0484}
        & \cellcolor{lampgreen}\textbf{3.0072} \\

        \midrule

        \multirow{4}{*}{ImageNet-R}
        & TENT
        & No
        & 41.84 & 0.5834 & 0.1649 & 3.2015 \\

        & \cellcolor{lampgreen}\MIRAmethod{TENT}
        & \cellcolor{lampgreen}No
        & \cellcolor{lampgreen}\textbf{44.10}
        & \cellcolor{lampgreen}0.5771
        & \cellcolor{lampgreen}\textbf{0.1361}
        & \cellcolor{lampgreen}\textbf{2.9875} \\

        & TENT
        & Yes
        & 40.74 & 0.5404 & 0.1330 & 3.2062 \\

        & \cellcolor{lampgreen}\MIRAmethod{TENT}
        & \cellcolor{lampgreen}Yes
        & \cellcolor{lampgreen}\textbf{41.13}
        & \cellcolor{lampgreen}0.5405
        & \cellcolor{lampgreen}\textbf{0.1292}
        & \cellcolor{lampgreen}\textbf{3.1631} \\

        \bottomrule
    \end{tabular}
\end{table}

MIRA improves accuracy, ECE, and NLL in both conventional and computationally
realistic settings. Although the gains become smaller under the realistic
protocol, they remain consistent across the evaluated metrics.

\subsection{Accuracy--Cost Trade-off}

Finally, we compare MIRA with TENT and computationally heavier TTA methods
under the same hardware setting.
Table~\ref{tab:accuracy_cost} reports classification performance, runtime,
and peak memory for the single-run CIFAR-100-C / ResNeXt comparison.

\begin{table}[H]
    \centering
    \small
    \caption{Single-run accuracy--cost comparison on CIFAR-100-C / ResNeXt (batch size 64). Bold marks the best values in directional columns.}
    \label{tab:accuracy_cost}
    \setlength{\tabcolsep}{5pt}
    \renewcommand{\arraystretch}{1.0}
    \MIRAcompacttable
    \begin{tabular}{lcccccc}
        \toprule
        \textbf{Method}
        & \textbf{Acc. (\%) $\uparrow$}
        & \textbf{MMC}
        & \textbf{ECE $\downarrow$}
        & \textbf{NLL $\downarrow$}
        & \textbf{Time (s) $\downarrow$}
        & \textbf{VRAM (MiB) $\downarrow$} \\
        \midrule

        TENT
        & 67.10
        & 0.7474
        & 0.0779
        & 1.2873
        & \textbf{125}
        & \textbf{2338} \\

        CoTTA
        & 64.16
        & 0.6170
        & \textbf{0.0529}
        & 1.4173
        & 1142
        & 3062 \\

        PETAL
        & 62.30
        & 0.5816
        & 0.0582
        & 1.5059
        & 1586
        & 3930 \\

        \rowcolor{lampgreen}
        \MIRAmethod{TENT}
        & \textbf{67.75}
        & 0.7310
        & 0.0556
        & \textbf{1.2401}
        & 157
        & 2494 \\

        \bottomrule
    \end{tabular}
\end{table}

Compared with TENT, MIRA introduces moderate additional runtime and memory
overhead while improving accuracy, ECE, and NLL. It remains substantially more
efficient than CoTTA and PETAL, supporting the practicality of cached
last-layer Monte Carlo sampling.

\subsection{Repeated Evaluation under Representation Shift}
\label{app:sketch_repeated}

ImageNet-Sketch~\cite{wang2019learning} tests a stronger representation shift.
Table~\ref{tab:sketch_repeated} uses ResNet-50, learning rate $2.5\times10^{-4}$,
and seeds 0, 1, and 2. MIRA improves mean accuracy by $1.35$ percentage points
and reduces ECE and NLL.

\begin{table}[H]
    \centering\small
    \caption{ImageNet-Sketch / ResNet-50, mean $\pm$ sample SD over seeds 0, 1, and 2. Bold marks best accuracy, ECE, and NLL.}
    \label{tab:sketch_repeated}
    \MIRAcompacttable
    \begingroup
\setlength{\tabcolsep}{4pt}
\renewcommand{\arraystretch}{1.0}
\MIRAcompacttable
\begin{tabular}{@{}lrrrr@{}}
\toprule
Method & Acc. (\%) $\uparrow$ & MMC & ECE $\downarrow$ & NLL $\downarrow$ \\
\midrule
Source & $23.84\MIRAsd{0.35}$ & $0.4623\MIRAsd{0.0013}$ & $0.2240\MIRAsd{0.0029}$ & $5.0889\MIRAsd{0.0089}$ \\
TENT & $28.22\MIRAsd{0.42}$ & $0.4507\MIRAsd{0.0031}$ & $0.1685\MIRAsd{0.0014}$ & $4.1317\MIRAsd{0.0163}$ \\
\rowcolor{lampgreen}
\MIRAmethod{TENT} & $\mathbf{29.57}\MIRAsd{0.29}$ & $0.4204\MIRAsd{0.0025}$ & $\mathbf{0.1247}\MIRAsd{0.0007}$ & $\mathbf{3.9175}\MIRAsd{0.0264}$ \\
\bottomrule
\end{tabular}
\endgroup

\end{table}

\subsection{Modified One-Step CLIP Evaluation}
\label{app:clip_one_step}

\begin{table}[H]
    \centering\small
    \caption{CLIP ViT-B/16 TENT-component comparison: one update per batch with a shared $10\times$ released learning rate. Bold compares each dataset pair.}
    \label{tab:clip_one_step}
    \MIRAcompacttable
    \begingroup
\setlength{\tabcolsep}{4pt}
\renewcommand{\arraystretch}{1.0}
\MIRAcompacttable
\begin{tabular}{@{}lrrrr@{}}
\toprule
Method & Acc. (\%) $\uparrow$ & MMC & ECE $\downarrow$ & NLL $\downarrow$ \\
\midrule
\multicolumn{5}{l}{\textbf{CIFAR-10-C}} \\
\midrule
TENT & 52.74 & 0.9443 & 0.4170 & 4.6682 \\
\rowcolor{lampgreen}
\MIRAmethod{TENT} & \textbf{54.13} & 0.9387 & \textbf{0.3977} & \textbf{4.5456} \\
\midrule
\multicolumn{5}{l}{\textbf{CIFAR-100-C}} \\
\midrule
TENT & 27.83 & 0.8382 & 0.5603 & 8.9705 \\
\rowcolor{lampgreen}
\MIRAmethod{TENT} & \textbf{32.71} & 0.8047 & \textbf{0.4784} & \textbf{7.5339} \\
\bottomrule
\end{tabular}
\endgroup

\end{table}

\section{Experimental and Implementation Details}
\label{app:app_D}

\subsection{Experimental Setup}
\label{app:setup}

We use corruption severity 5 unless otherwise stated. Source training,
preprocessing, and target streams are matched within each base/MIRA comparison.
All experiments are conducted on NVIDIA RTX A6000 GPUs.

For the TENT pairs in Table~\ref{tab:plug_and_play_multi_model}, both benchmarks
use 15 severity-5 corruptions, batch size 64, and 10,000 shuffled examples per
corruption. Model and optimizer are reset between corruptions, and each
base/MIRA pair receives the same ordered streams. MIRA uses the feature-wise
He-in construction in Appendix~\ref{app:scale} and the defaults in
Appendix~\ref{app:hparams}.

The loss, construction, and sensitivity comparisons use paired seeds 1, 2,
and 3. Scope, budget, architecture diagnostics, and most stream/ViT/CLIP
extensions are single-run.

\subsection{Evaluation Metrics}
\label{app:metrics}

We report classification accuracy, mean maximum confidence (MMC), expected
calibration error (ECE), and negative log-likelihood (NLL).

Let $p_i$ denote the predictive probability for sample $i$,
$\hat{y}_i=\arg\max_c p_{i,c}$, and $c_i=\max_c p_{i,c}$.
Accuracy is the fraction of correct top-1 predictions, while
$\mathrm{MMC}=\frac{1}{N}\sum_i c_i$ measures average predictive confidence.

ECE uses 10 equal-width confidence bins~\cite{guo2017calibration}, and NLL
uses the probability assigned to the correct class:
\begin{equation}
\begin{aligned}
\mathrm{ECE}&=\sum_{m=1}^{10}\frac{|\mathcal B_m|}{N}
\left|\mathrm{acc}(\mathcal B_m)-\mathrm{conf}(\mathcal B_m)\right|,\\
\mathrm{NLL}&=-\frac{1}{N}\sum_{i=1}^{N}\log p_{i,y_i},
\end{aligned}
\end{equation}
with probabilities lower-bounded by $10^{-12}$ for numerical stability.
Each metric is computed over the evaluated examples of each corruption and
then macro-averaged across corruptions.

For three-run results, we report sample standard deviation (SD),
$s=\sqrt{\sum_{r=1}^{3}(v_r-\bar v)^2/2}$, rather than the standard error
$s/\sqrt{3}$. Table~\ref{tab:loss_ablation} reports accuracy and its SD as
fractions; accuracy columns labeled (\%) use percentages, with SD in
percentage points. MMC, ECE, and NLL use their native units. Run variability
includes target-stream order, batch composition, Gaussian sampling, and cuDNN
nondeterminism. SD describes variability rather than establishing statistical
significance.

\subsection{Baseline Implementation Details}

We retain each base method's optimizer settings.
For the TENT pairs in Table~\ref{tab:plug_and_play_multi_model}, SGD uses a
learning rate of $10^{-3}$ on CIFAR-100-C / ResNeXt and $2.5\times10^{-4}$
on ImageNet-C / ResNet-50, with seeds 1, 2, and 3 on both benchmarks.
CIFAR-100-C EATA and ROID use a learning rate of $2.5\times10^{-4}$ and seeds
1, 2, and 3; their ImageNet-C counterparts use seeds 2, 4, and 6.
We use momentum $0.9$ and zero weight decay unless
otherwise specified. When MIRA is integrated with a base TTA method, we retain
the optimizer and learning rate of that method rather than retuning them.

The ImageNet-C / ResNet-34 loss ablation in Table~\ref{tab:loss_ablation}
uses SGD at $2.5\times10^{-4}$, $K=300$, 15 severity-5 corruptions,
and reset between corruptions.

For EATA, we additionally use Fisher regularization of $1.0$, with the
Fisher information matrix estimated from 2000 source samples. Fisher estimation
is a source-dependent component of the EATA baseline; attaching MIRA introduces
no additional source-data requirement.

We also compare against uncertainty-estimation baselines including Deep
Ensemble (DE), LLLA, ICLA, BBB, and ABNN. DE averages predictions from
independently trained source models. LLLA follows the official Laplace
implementation, while the remaining methods follow their released
implementations or published settings using the same backbone and evaluation
protocol whenever possible.

\subsection{Perturbation Scale Construction}
\label{app:scale}

Let $W\in\mathbb{R}^{C\times D}$ denote the final classifier weight and
$h_i\in\mathbb{R}^{D}$ the cached feature of target sample $i$.
All headline experiments use a He-style base scale with a feature-wise,
batch-dependent factor:
\begin{equation}
\begin{aligned}
s_0&=\sqrt{2/D}, & r_{\mathcal B,d}&=\frac{1}{B}\sum_i |h_{i,d}|,\\
g_{\mathcal B,d}&=\mathrm{clip}\!\left(
\frac{r_{\mathcal B,d}}{D^{-1}\sum_j r_{\mathcal B,j}},0.25,4.0\right),\\
A_{\mathcal B}[c,d]&=s_0 g_{\mathcal B,d}, & a_{\mathcal B}[c]&=s_0,
\end{aligned}
\end{equation}
where $g_{\mathcal{B},d}$ is shared across output classes.
The He-style factor accounts for classifier input dimension, while the
feature-wise factor adapts the perturbation geometry to the current target
representation. Neither factor is tuned per dataset or model; only the global
perturbation radius $\sigma$ is selected online.

\subsection{Hyperparameter Settings}
\label{app:hparams}

Full MIRA uses the common defaults
\begin{equation}
\alpha=0.03,\quad\rho=0.90,\quad\tau=0.25,\quad\gamma=1,
\quad K=300,\quad B=64.
\end{equation}
For all headline experiments, we use the fixed candidate scale set
\begin{equation}
\begin{aligned}
\mathcal{G}={}&\{
10^{-5},
10^{-4},
3{\times}10^{-4},
10^{-3},
3{\times}10^{-3},
10^{-2},
3{\times}10^{-2},
5{\times}10^{-2},\\
&\quad 10^{-1},
3{\times}10^{-1},
5{\times}10^{-1},
1,2,3,4,5\}.
\end{aligned}
\end{equation}

The same Monte Carlo noise samples are reused across candidate scales within
each target batch. Since the backbone features are cached, scale evaluation
requires only repeated final-layer operations.

No corruption-specific or dataset-specific tuning is performed.

\subsection{Scale Selection}
\label{app:selector}

At each grid candidate we compute batch-average MI and agreement using shared
Gaussian draws. The implementation forms
$T_M=\min_j M(\sigma_j)+\alpha[\max_j M(\sigma_j)-\min_j M(\sigma_j)]$
and selects the first candidate satisfying $M\ge T_M$ and $A\ge\rho$.
When the preceding candidate lies below $T_M$ and both interval endpoints meet
the agreement condition, MI is interpolated linearly along $\log\sigma$.
If no candidate satisfies both conditions, $\sigma^*$ is the first candidate
satisfying the MI condition, which always exists for $\alpha\le 1$; the disagreement bound does not apply to such selections. The MI-range minimum need not equal the first candidate's MI in a numerically
non-monotone curve. After selection, independent Gaussian draws are used for
the loss and prediction at $\sigma^*$.

Default16 is the candidate set specified above.
Coarse9 is
\begin{equation}
\{10^{-5},3\times10^{-4},3\times10^{-3},3\times10^{-2},10^{-1},
5\times10^{-1},2,4,5\}.
\end{equation}
Dense31 inserts the geometric midpoint between each pair of default candidates;
Extended20 adds $\{10^{-6},3\times10^{-6},10,20\}$ to Default16.
All grid comparisons keep $K=300$ and identical common-noise shape.

For deployment, increase $\alpha$ when probes repeatedly remain weak, and reduce
$\alpha$ when perturbations frequently change class predictions. The agreement
bound applies only to selections that satisfy both conditions. Appendix~B.5
reports the joint sweep.

\subsection{CLIP Implementation and Configuration}
\label{app:clip}

CLIP ViT-B/16 uses the released CLIPTTA~\cite{lafon2025cliptta} architecture, data pipeline, and TENT
component. Logits are similarities between normalized image features and
class-specific text embeddings. The text-embedding matrix is treated as the
effective classifier and receives Gaussian perturbations only for reliability
probing; the unperturbed embeddings remain fixed. Adaptation updates visual
encoder normalization parameters, not the text embeddings.

Table~\ref{tab:vit_clip_generalization} uses the released ten-update-per-batch
TENT-component configuration, with identical optimization settings for TENT and
MIRA and no MIRA-specific tuning. The modified comparison in
Table~\ref{tab:clip_one_step} uses one update and a shared $10\times$ learning-rate
multiplier. The effective similarity classifier is not a learned linear head, and these
comparisons assess the TENT component rather than the complete CLIPTTA method.

\Needspace{380pt}
\section{Reliability Analysis}
\label{app:app_E}
\label{app:epistemic_aleatoric}

\begin{figure}[H]
\centering
\includegraphics[width=\linewidth]{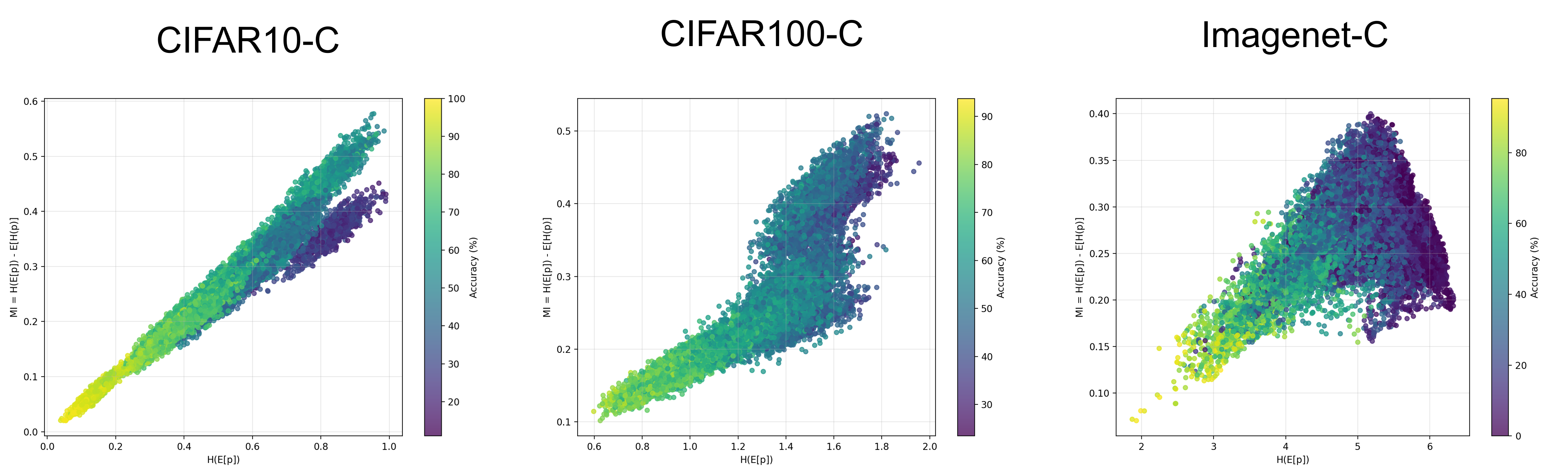}
\caption{Predictive entropy and perturbation-induced MI across CIFAR-10-C,
CIFAR-100-C, and ImageNet-C (left to right); color denotes batch accuracy.
MI measures local classifier disagreement, rather than calibrated Bayesian uncertainty.}
\label{fig:supp_he_eh}
\vspace{-6pt}
\end{figure}

Across CIFAR-10-C, CIFAR-100-C, and ImageNet-C, Figure~\ref{fig:supp_he_eh}
relates predictive entropy and perturbation-induced MI to batch accuracy.
These empirical diagnostics complement the fixed-scale example in Figure~\ref{fig:main_results}(b).

\begin{wrapfigure}{r}{0.55\textwidth}
\vspace{-6pt}
\centering
\includegraphics[width=\linewidth]{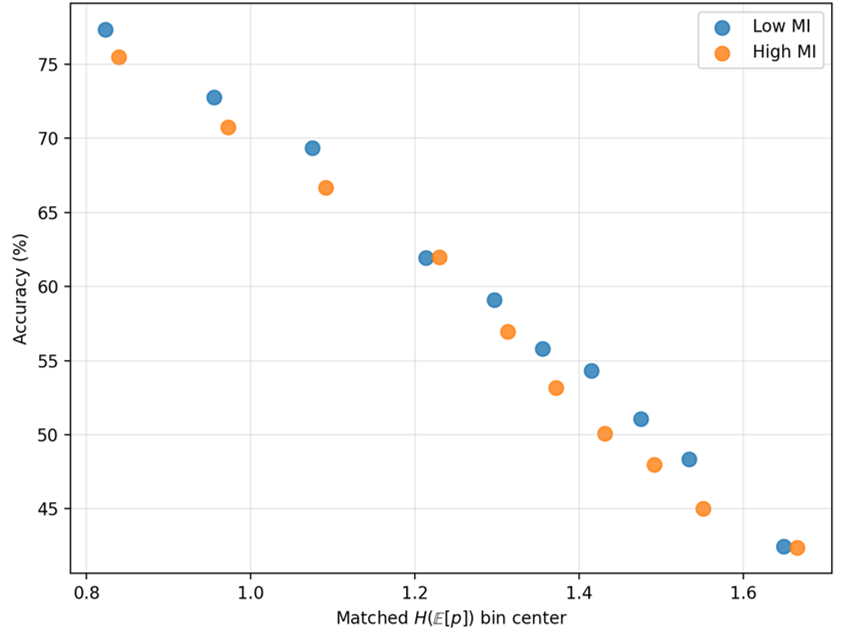}
\caption{Batch accuracy at matched predictive-entropy bin centers, pooling
corruption types. Low-MI batches tend to be more accurate than high-MI batches.}
\label{fig:mi_base_appendix}
\vspace{-6pt}
\end{wrapfigure}

To test whether MI provides information beyond predictive entropy, we compare
batches with similar $H(\mathbb E[p])$. Figure~\ref{fig:mi_base_appendix} shows
that high-MI batches tend to have lower accuracy than low-MI batches at similar
entropy. Thus, local perturbation sensitivity supplies a complementary
reliability signal. This observational comparison supports the interpretation
of MI and does not establish a causal benefit from weighting.

Matching predictive entropy makes the two groups comparable in the overall
sharpness of their mean predictions. At a similar bin center, the remaining
accuracy gap therefore reflects an association with perturbation sensitivity
beyond the mean entropy alone. The groups approach one another in some bins,
so MI is a relative reliability cue within the probed classifier neighborhood,
rather than a calibrated error probability. This is consistent with MIRA's
detached weighting: low-MI samples receive more influence, while the confidence
term separately limits updates from low-evidence predictions. The analysis
therefore motivates combining the two signals instead of treating either as
a complete description of reliability.

\end{document}